# Clinical Document Corpora – Real Ones, Translated and Synthetic Substitutes, and Assorted Domain Proxies: A Survey of Diversity in Corpus Design, with Focus on German Text Data

## Udo Hahn

Institute for Medical Informatics, Statistics and Epidemiology (IMISE), University of Leipzig, Leipzig, Germany

hahn@texknowlogy.com

**ABSTRACT**

**Objective:** We survey clinical document corpora, with focus on German textual data. Due to rigid data privacy legislation in Germany these resources, with only few exceptions, are stored in protected clinical data spaces and locked against clinic-external researchers. This situation stands in stark contrast with established workflows in the field of natural language processing where easy accessibility and reuse of (textual) data collections are common practice. Hence, alternative corpus designs have been examined to escape from this data poverty. Besides machine translation of English clinical datasets and the generation of synthetic corpora with fictitious clinical contents, several other types of domain proxies have come up as substitutes for real clinical documents. Common instances of close proxies are medical journal publications, clinical therapy guidelines, drug labels, etc., more distant proxies include medical contents from social media channels or online encyclopedic medical articles.

**Methods:** We follow the PRISM (Preferred Reporting Items for Systematic reviews and Meta-analyses) guidelines for surveying the field of German-language clinical/medical corpora. Four bibliographic databases were searched: PubMed, ACL Anthology, Google Scholar, and the author's personal literature database.

**Results:** After PRISM-conformant identification of 362 hits from the four bibliographic systems, after the screening process 78 relevant documents were finally selected for this review. They contained overall 92 different published versions of corpora from which 71 were truly unique in terms of their underlying document sets. Out of these, the majority were clinical corpora – 46 real ones from which 32 were unique, 5 translated ones (3 unique), and 6 synthetic ones (3 unique). As to domain proxies, we identified 18 close ones (16 unique) and 17 distant ones (all of them unique).

**Discussion:** There is a clear divide between the large number of non-accessible authentic clinical German-language corpora and their publicly accessible substitutes: translated or synthetic datasets, close or more distant proxies. So, at first sight, the data bottleneck seems broken. Intuitively yet, differences in genre-specific writing style, wording and medical background expertise in this typological space are also obvious. This raises the question how valid alternative corpus designs really are. A systematic, empirically grounded yardstick for comparing real clinical corpora with those suggested substitutes is missing up until now.

**Key words:** Natural language processing, Clinical text corpora, Medical text corpora, German language

## LAY SUMMARY

Corpora, i.e., collections of textual, audio or visual data, are crucial for training and evaluating language models which are the backbone of down-stream application tasks, such as information extraction, text mining, or document classification. Due to ethical concerns and corresponding legislation, access to clinical corpora is severely restricted world-wide. Particularly high distribution hurdles have been implemented in Non-Anglo-American regions of the world, especially Europe. To illustrate this corpus dilemma we focus on the current situation Germany. We review in depth real, i.e., authentic German-language clinical corpora and then, due to their prohibitive access conditions, widen our scope to corpus design alternatives to break this data bottleneck. Several substitutional approaches have been pursued, such as translations from English clinical datasets to German, the construction of synthetic corpora with fictitious contents, and close as well as more distant domain proxies. The latter two incorporate documents with medical themes yet feature entirely different text genres and writing styles, such as medical journal articles, clinical therapy guidelines, or drug labels as close domain proxies, and medical social media contents as well as online encyclopedic medical articles as more distant domain proxies. Unlike real clinical corpora, almost all these potential substitutes are publicly available and, thus, alleviate data sparsity. An open empirical research question remains though: at what costs (e.g., in terms of system performance) can these alternative corpus designs substitute real clinical documents?

**BACKGROUND**

Corpora are collections of so-called *unstructured* textual, audio or visual data in contrast to structured, mostly tabular, information stored in databases or spreadsheets. Whereas structured data is readily interpretable and thus actionable by computers, unstructured data is not. To computationally interpret unstructured data language models are automatically learned which capture and represent the data's structure and contents so that computers can reason on the models' representation structures. This learning process is either organized in an unsupervised way, just relying on typically huge masses of raw data and the distributional patterns they embody, or in a (semi-)supervised manner where metadata explicitly inform the machine learning engine with crucial (syntactic and) semantic interpretation hints.

Typically, such metadata are supplied by humans as the result of annotation processes that lead to *gold standard* data (so-called ground truth); automatic tagging may replace humans in the loop and yields (typically, lower quality) machine-generated annotations, a computational process that generates *silver standard* data. Annotations mimic the human understanding process of unstructured data by requiring human annotators to strictly follow interpretation rules laid down in carefully crafted annotation guidelines. The outcome of annotation processes is quality-checked in terms of inter-annotator agreement (IAA) metrics whose scores indicate how close annotators adhere to annotation guidelines as language understanders (for comprehensive surveys on the role of corpora for machine learning and natural language processing, see [1,2]; for an introduction to the organization of and methodology underlying annotation campaigns, see [3,4]). Annotated corpora are typically built with specific purposes in mind, e.g., down-stream applications such as text classification, named entity recognition or relation/event extraction. Consequently, they normally address only one specific target layer of (language) understanding rather than its whole multi-dimensional spectrum.

Over the years, corpora have turned into an indispensable prerequisite for natural language processing (NLP) since they serve two purposes. First, they provide the input for *machine learning* algorithms to learn structural and content properties from unstructured data. Second, annotated corpora constitute a common ground for evaluation experiments to measure the quality of systems operating on unstructured data in terms of (community-consensual) *benchmarks*. Hence, well-designed, reasonably sized and publicly shared corpora are the foundation for the *reproducibility* of research results in that they allow the solid comparison of different types of language models, different sets of (hyper)parameters within the same model family, their effect on the outcomes of down-stream tasks, or alternative system architectures, etc.

The dire need for specialized *clinical* corpora arises from the fact that medicine, as many other sciences, has established a highly diversified sublanguage on its own, diverging strongly from other scientific disciplines beyond the life sciences and, in particular, common language use patterns in every-day verbal communication [5,6]. Even worse, clinical language is not homogeneous but splits into numerous subdomains and text genres [7,8,9] also differing from each other in many ways. Therefore, the utility of a given clinical/medical corpus must be carefully assessed in the light of various descriptive dimensions:

- Medical *subdomains* are often incompatible at the terminological level and follow different reporting standards. Consequently, documents from oncology are different from cardiology or radiology, and vice versa, both in terms of document structure and the verbalization of contents. This raises the issue whether a multitude of homogeneous *subdomain corpora* have to be supplied as an adequate pool for model training or, when such a large spectrum of subdomain corpora is lacking, whether models trained on, say, oncology data lead to poor(er) cardiology or radiology models, and vice versa.
  Liang *et al.* [10], e.g., report on domain transfer learning experiments where PubMed-based generic medical language models (derived from medical journal abstracts) are applied to oncology data (the Bronco corpus [11]) and nephrology data (the Ex4CDS corpus [12]), respectively. Their results yield preliminary evidence that much of the enormous variance in the classification results can be attributed to the semantic alignment of (merged) named entity types to harmonize the corpora involved. In a follow-up study [13], the authors tackle this problem of semantic diversity by proposing a multi-layered semantic annotation scheme.
- Similar discrepancies can be observed for different clinical *text genres*.[1] Discharge summaries differ significantly from pathology reports, radiology reports, operative reports, or nursing notes, both in terms of document structure and the verbalization of contents. Again, the question pops up whether homogeneous *genre-specific corpora* are needed for model training or, put the other way round, whether models trained on, say, discharge summary data lead to poor(er) pathology or radiology report models, and vice versa.
- Another crucial source of variance relates to *site-specific documentation standards*. For instance, discharge summaries from clinic A may deviate from those produced in clinic B and C, both in terms of document structure and verbal realization. This raises the question whether homogeneous *site-specific corpora* have to be generated for proper model training (even for the same clinical domain and clinical report genre) or, phrased alternatively, whether models trained on discharge summaries from clinic A are valid, at all, for discharge summaries from clinic B or C, and vice versa.

  Böhringer *et al.* [14] conducted experiments to automatically infer ICD-10 codes in ophthalmologic departments of three different German hospitals and found that common eye disorders were mostly accurately classified by a language model trained in one of these hospitals and rolled out in the other two hospitals whereas others, rare diseases in particular, varied considerably in classification accuracy. The authors also noted diverging local terminological standards and reporting habits, e.g., the use of uncommon abbreviations that only make sense and are only understood in the local hospital environment. Such local "dialects" add an additional level of complexity to corpus building initiatives hard to cope with.

Perhaps the most problematic issue with clinical/medical corpora is tied to the quest for prioritizing individual data privacy and security over distributability which leads to extremely high hurdles, if not a non-negotiable blockade, to make such corpora publicly available. The underlying ethical concerns [15] have been translated into legal protection regulations world-wide. These are intended to avoid individual patients' re-identification once clinical documents leave safe, hospital-internal data spaces, such as the patients' Electronic Health Record (EHR). Criteria deserving such protection efforts have been spelled out most explicitly in the US HIPAA legislation act[2] and cover 18 privacy-sensitive attributes, so-called *Personally Identifiable Information* (PII),[3] which carry information about the patients' and other clinical actors' identity (see, e.g., Table 1 in [16]). HIPAA's safe harbor rules require that such data items be neutralized by de-identification processes prior to allowing use by or disclosure to clinic-external individuals or institutions. Under these provisions data distribution allowance usually requires signing a *Data Use Agreement* (DUA) between data owners and external data users which spells out detailed protective conditions for data storage and use at external sites. In Europe, the conditions of the *General Data Protection Regulation* (GDPR)[4] and national Germany data protection laws (e.g., *"Gesundheitsdatennutzungsgesetz"* (GDNG))[5] are less explicit in that they lack a comparable list of attributes. Instead, they are even more restrictive requiring the explicit informed consent of data subjects for any external use. In essence, these requirements have the following implications:

- Clinical corpora may, under no circumstances, be made publicly available without complete and certified de-identification of PIIs. Certification and clearance are usually administered by the ethical board of the local hospital the data come from.
- The features or attributes to be identified are explicitly enumerated for the US clinical NLP community (HIPAA's PIIs). Such a clarification is missing in European law (GDPR) and national German law (GDNG). GDPR posits that data subjects have to express explicit consent that their de-identified data can be used for subsequent information processing, German law vaguely states that de-identified data can only be made publicly available when privacy can be broken with "unreasonable efforts" (what unreasonable efforts really are is not spelled out).
- The allowance for (fully de-identified) clinical corpora to be publicly distributed is always bound to the consent of the ethics board of the local hospital the corpus emerged from. This decision is based on verified adherence to the current legal data security ecosystem in Germany, as well as local hospital rules and practices. Clinical administrations in Germany are extremely cautious to avoid potential juridical measures against data clearance and thus usually block corpus distribution.

[1] The diversity of text genres is truly amazing. A *de facto* standard for the categorization of clinical documents in Germany, *Klinische Dokumentenklassen-Liste* (KDL), distinguishes more than 400 different genre types (https://simplifier.net/kdl/kdl-cs-2025). We owe this information to Frank Meineke (personal communication).

Given this legal frame of reference, only very few German-language clinical corpora have been released for public use up until now. Consequently, the clinical NLP community in Germany has made immense efforts to replace real clinical corpora by reasonable substitutes or domain proxies. All these efforts are documented in detail in the Supplementary Material section of this article and will be summarized in the Results section.

---

[2] https://www.hhs.gov/hipaa/index.html
[3] https://www.directives.doe.gov/terms_definitions/personally-identifiable-information-pii
[4] https://gdpr.eu/
[5] For a detailed discussion of German data protection regulations pertaining to clinical corpus distribution, see Lohr *et al.* [17, Section 3].

## OBJECTIVE

This review sheds light on corpus developments in the clinical, and more broadly medical, domain for the German language (spoken primarily in Germany, Austria and parts of Switzerland by roughly 100 million native speakers). We will report on various real clinical corpora almost all of them locked in safeguarded clinical data silos. Due to legal privacy protection regulations in Germany clinic-external distribution of these corpora is usually forbidden even after strict HIPAA-style de-identification so that they remain inaccessible to the wider (clinical/medical) NLP community. Such rigid access restrictions violate established routines in NLP R&D workflows in which the (re-)usability of corpora is common practice for training and evaluating language models. Corpus developers have thus investigated several alternatives to bypass this data bottleneck. Hence, we will also review these potential substitutes for real clinical corpora in depth (for alternative surveys of German clinical corpora, see [18,19]).

This review targets the following objectives:

- We provide a comprehensive survey of *German-language* corpora in the *clinical* domain and complement this narrow view by corpora with a wider *medical* scope.
- The corpora included in this review deal with *written* verbal data (only). As far as multi-media data (e.g., images in radiology reports) are concerned, only the written portion is dealt with. Speech corpora with spoken language as primary verbal data (e.g., audio records of doctor-patient conversations) and any other modality complementing language behavior (visual information via deictic pointing gestures, body movements, facial expressions, etc.) will be excluded from this survey.
- We cover (hopefully) all corpora which have been published under peer review policy in the past quarter of a century, namely from *2000 until December 2024*.
- Abstracting away from the specifics of the individual corpora we survey, we introduce a generic template, we call *corpus card*, to guide future corpus descriptions (see Appendix A). This recommendation is language-independent and may be useful, in general, for the international medical informatics community to promote higher data science standards for corpus documentation.

## MATERIALS AND METHODS

We followed the PRISM (Preferred Reporting Items for Systematic reviews and Meta-analyses) guidelines [20] for surveying the field of German-language clinical/medical corpora.

**Study Identification.** Since the topic of this review lies at the intersection of (clinical) medicine and NLP, we considered a medical bibliographic resource (PubMed® which comprises more than 37 million citations for biomedical literature from the bibliographic database Medline) and an NLP-focused one (ACL Anthology, with up to 100,000 bibliographic units from the most authoritative institution in the field of NLP, the *Association for Computational Linguistics*). As a third resource, we took Google Scholar (whose focus is on thematically unconstrained scholarly publications). Finally, the author's own bibliographic database, ABIB (with more than 65,000 bibliographic units covering (biomedical) NLP publications), was searched as well. The following queries were evaluated on August 24, 2024, on all four bibliographic databases (in addition, we conducted a final search on ABIB on January 15, 2025, to collect the latest publications from 2024):

PUBMED

Query: **(german) AND (text OR document) AND (corpus)**
Hits: **89**

ACL ANTHOLOGY

Query: **(german) AND (clinical OR medical) AND (corpus)**
Hits: **5,510** (ordered by relevance)

GOOGLE SCHOLAR

Query: **(german) AND (clinical OR medical) AND (corpus)**
Hits: **~ 443.000** (ordered by relevance)

ABIB

Query: **(language: german) AND (domain: medicine OR domain: clinic) AND (text corpus)**
Hits: **70 (+3) = 73**

All hits were checked for PUBMED (89) and ABIB (73) whereas only the first 100 hits could be screened for ACL (the list was truncated after 100 hits by the search engine and could not be expanded) and GOOGLE (to mimic the procedure for ACL). The PRISM flowchart for the document selection process is depicted in Fig. 1, while the distribution of all relevant articles and their overlaps for the four different search engines are displayed in Fig. 2.

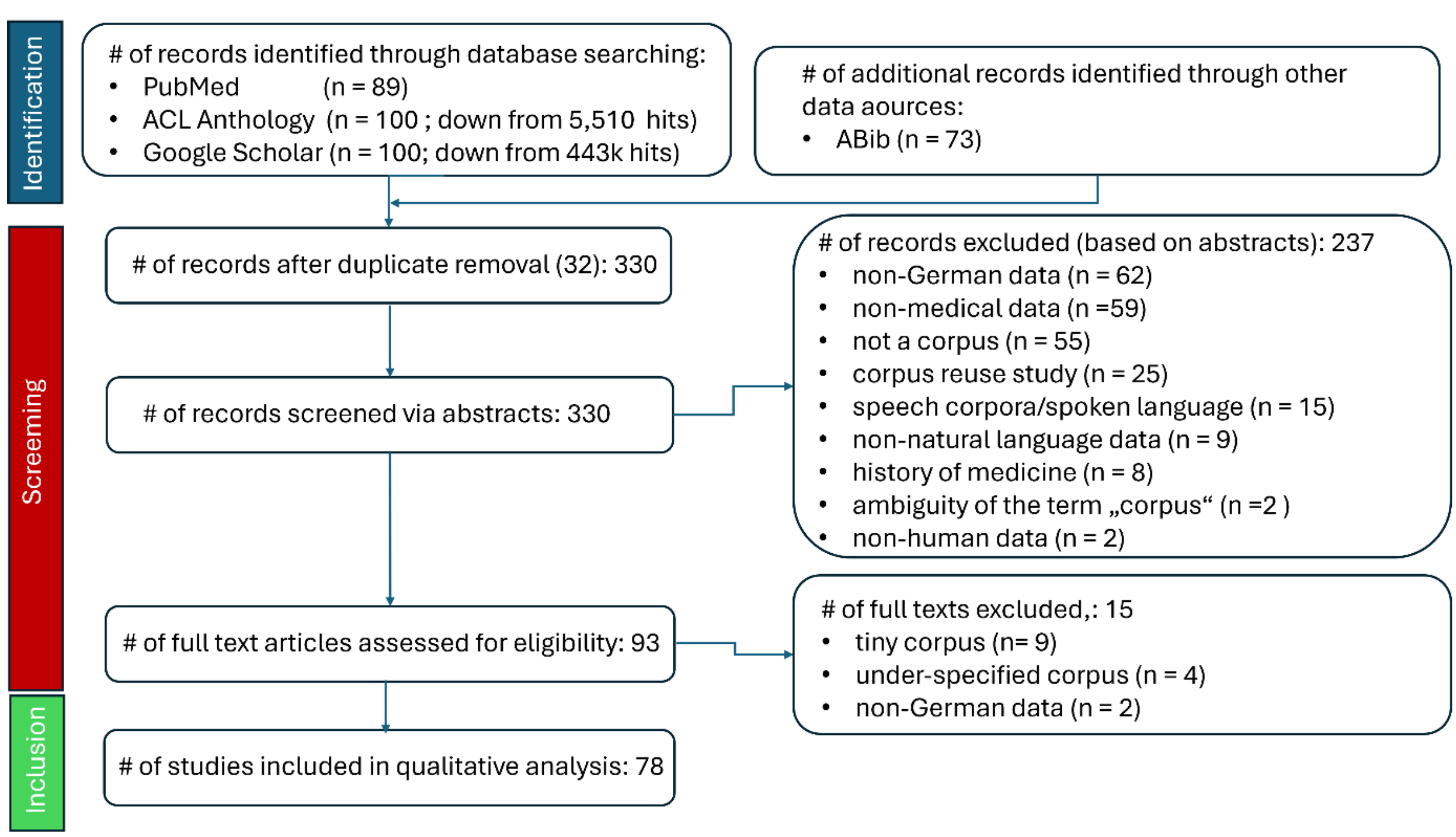

**Figure 1:** PRISM Flowchart

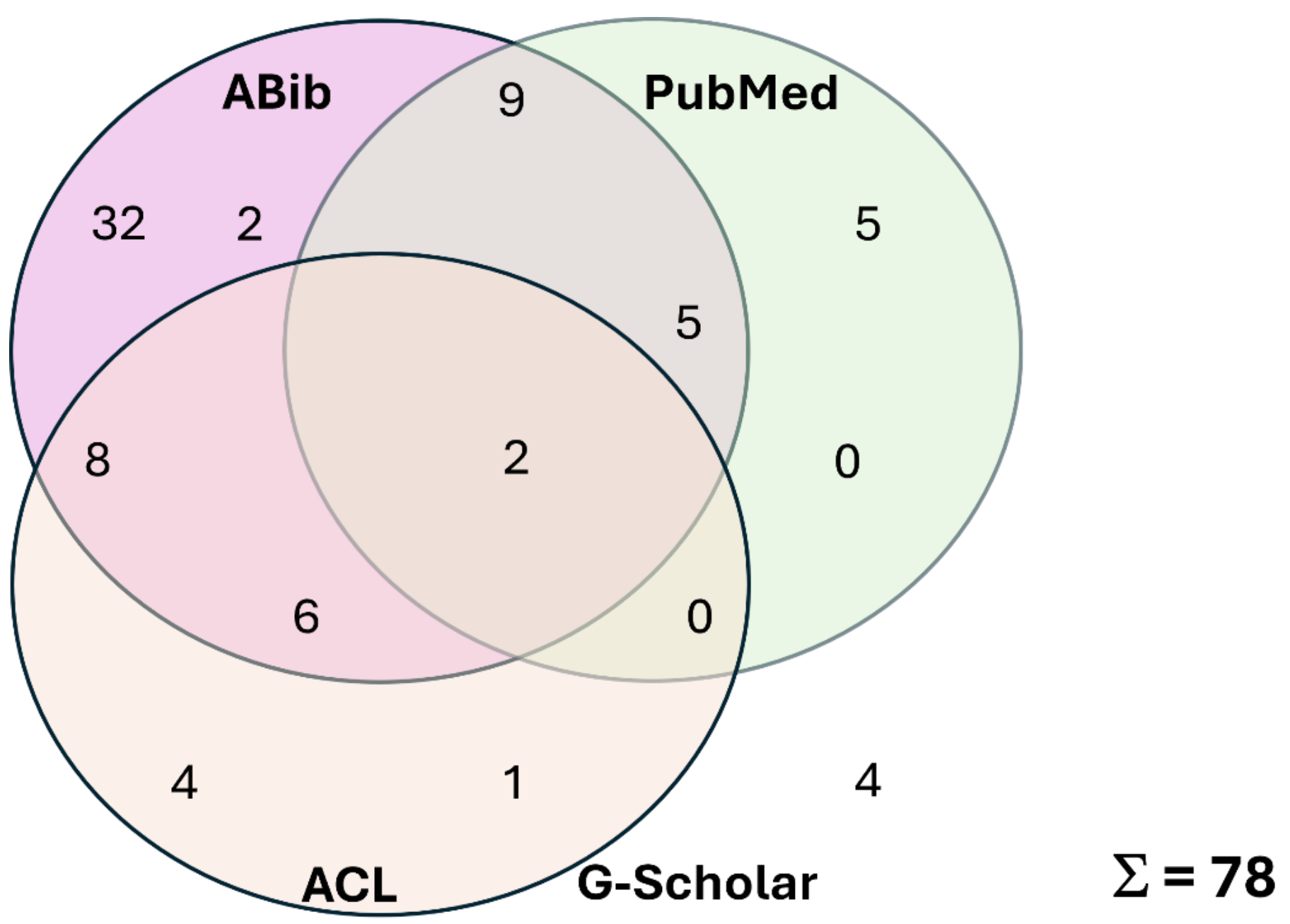

**Figure 2:** Distribution and Overlap of Relevant Hits

**Eligibility criteria**. Only German-language clinical/medical corpora were eligible for this review; mixed-language corpora (e.g., parallel corpora) were included if they contained a significant German portion (see criteria below). Publications that reused already existing corpora for down-stream applications were excluded, as well as corpora featuring *spoken* language, i.e., audio data, whereas *written* chats, blogs, and tweets from social media channels or *written* doctor-patient conversations were included. Tiny corpora with less than 100 documents or less than 10,000 tokens were discarded (unless they are publicly shareable), as well as corpora portraying the history of medicine. Overly under-documented corpora lacking fundamental descriptive data (e.g., number of documents or tokens) were also eliminated. We focused on human medicine only. The four independent searches yielded 362 hits altogether from which 78 were considered relevant and, thus, form the basis for this review.

## RESULTS

The following presentation of results is based on the division of corpus descriptions into five tables that can be found in the Supplementary Material (see Tables 1 to 5). We distinguish between three types of clinical corpora (namely, real or authentic, translated, and synthetic ones) and two types of non-clinical, medical corpora as domain proxies (mainly built from scholarly medical publications on the one hand, and social media data and encyclopedic articles, on the other hand). For all five categories of corpora, we distinguish between

- the number of *publications* in which the individual corpora are described per category,
- the number of *distinct* (or *document-unique*) corpora per category, i.e., ones with zero intersection of their document sets, or, alternatively, where different versions of the same corpus are genealogically aligned (this criterion merges identical document sets or sets of documents where one corpus is a superset of another one), and
- the number of *annotation-unique* corpora per category, i.e., ones to which different types of metadata have been assigned (corpora lacking any metadata are excluded).

A summary table of all German-language clinical/medical corpora in which these distinctions will be made concrete appears in the Discussion section (see Table 6).

### Clinical Corpora

***Real Clinical Corpora***.

Real clinical corpora are composed of original clinical reports or notes written by professional clinical staff who report about individual patients during their hospital stay. We found 46 publications for such corpora from which 32 are distinct (document-unique) whereas 40 corpora are annotation-unique, i.e., annotated with different types of metadata. Table 1 in the Supplementary Material section gives a detailed overview of these 46 corpora.

Clinical corpus construction efforts for the German language started in 2004 with FraMed [21]. This corpus is small-sized (100k tokens), annotated with low-level linguistic information only, and (due to the inclusion of clinical and copyrighted textbook material) non-sharable as a dataset. Yet, language models for sentence and token splitting as well as part-of-speech tagging were made publicly available in the JCoRe model release ten years later [22,23]. From 2007 to 2016 various clinical corpora were developed as a by-product of application-focused studies, with Müller-07 [24], Kreuzthaler-11 [26] and Bretschneider-14 [29] constituting, at that time, quantitatively outstanding datasets (roughly 30,000 documents (no token count), 3,500 documents, 84k tokens, and 2,700 documents, 347k tokens, respectively); Müller-07 and Kreuzthaler-11 come without any medically relevant metadata, whereas Bretschneider-14 has 148k tokens semantically annotated with domain-specific RadLex terms for radiology reports.

Around 2015, several new tendencies can be observed for corpus building in the German-language clinical NLP community. First, corpora, once created, undergo continuous quantitative augmentation, qualitative curation and, in general, profit from iterative refinement in follow-up studies. Furthermore, the annotations feature fine-grained semantic information in terms of clinically relevant named entity and semantic relation types, as well as linguistic information covering, e.g., negation and uncertainty signals. A typical example of this move are the activities of the Roller group [33,35,37,47,53] who developed a homogeneous corpus of discharge summaries in the nephrology domain (about 1,725 (1,360) documents, with some 158k (111k) tokens). It excels in the richest semantic type repertoire up until now (around 46k named entity annotations for 17 types and 17k relation annotations for 9 types in the latest, slightly downsized release [53]). For the first time ever, also a DUA-based access option for pre-trained information extraction models is provided. The approach taken by the 3000PA team [39,40,42,45,60] is perhaps even more ambitious, since their work (based on more than 6,600 clinical documents, mainly discharge summaries, from three different national university hospitals, with 7,3 million tokens [60]) aims at the broad coverage of very diverse annotations layers ranging from medication information (1 entity type, 5 relations) [39], 18 section heading types [40], 13 PII entity types [42], 3 medical named entity types (Symptom, Finding, Diagnosis) [45], various semantic relations, as well as factuality and temporal information [60]. All this accumulates in slightly more than 2 million annotation items in the final release [60], a metadata resource unmatched in quantity and breadth. Work on Cardio:DE [56] (formerly named CardioAnno [49]) features 500 clinical reports (993k tokens) in the cardiology domain [56], with 12 cardiovascular entity types (1,6k annotation units) [49], 14 section heading types (116,9k annotation units), 2 named entity types for medication and 7 relation types (26,6k annotation units) [56]. Unlike the previously built corpora, Cardio:DE is publicly available on a DUA basis. Finally, the work of Bressem *et al.* features 6,000 radiology reports (estimated 850k tokens), with annotations relating to 9 Finding types (15k annotation units) [46], the presence/absence of 4 pathologies, and 4 different types of therapy devices [61]. The radiology core of this corpus remains locked, yet the RadBert language model for extracting Finding types [46], as well as pretrained model weights for the medBert language model and radiology benchmarks can be distributed [61]. These studies, fully compliant with mainstream non-medical NLP, also mark a fundamental change of the role of corpora in clinical NLP – originally conceived as a side issue of application-centered research their design and realization now has become a respected research theme on its own.

When judging the potential value of clinical corpora quantity in terms of the number of documents or tokens is only a weak indicator. For instance, the largest corpora in terms of the number of documents, IDRISSI-YAGHIR-24 [62], with slightly more than 25,000k documents, MEDCORPINN [50,51], with 5,000k documents, GRUNDEL-21 [48], with 40,5k documents, and OLEYNIK-17 [36], with 30k documents, all suffer from the lack of any clinically relevant metadata (GRUNDEL-21 inherits gold standard data from the structured part of the parallel EHR from which the documents were extracted). Within this group of very large corpora, only DMP "HERZMOBIL" [63], with roughly 36k documents, carries medically relevant semantic annotations, yet these are automatically generated and thus form a silver standard corpus. Also, due to the nature of different clinical document genres (e.g., discharge summaries being much longer than clinical notes), the number of tokens does not necessarily increase with the number of documents. As an alternative yardstick for content-based corpus assessment one might prefer the numbers of semantically rich, medically relevant annotations. On that scale, the following corpora are top-ranked:

- 3000PA 5.0 [60], with 6,600 documents (7,300k tokens), composed of discharge summaries, with 2,093k multi-level annotation units,
- CARDIO:DE [56], with 500 documents (993k tokens), composed of clinical reports from the cardiology domain, with 143,5k named entity and relation annotations,
- ROLLER-20 [47], with 1,725 documents (158k tokens), composed of discharge summaries from the nephrology domain, with 77,4k named entity and relation annotations.

Sheer numbers relating to documents, tokens, and medical metadata are but one side of the coin for corpus assessment. On the flipside, their accessibility to a wider R&D community is even more important for scientific progress. Here comes the bad news – out of 32 document-unique corpora, only 5 are externally accessible at all, yet with different clearance policies. A historical breakthrough was achieved with BRONCO [11], a collection of 200 discharge summaries (90k tokens), with annotations for section headings and 3 named entity types, Diagnosis, Treatment, and Medication, plus their grounding in ICD-10, OPS, ATC terminologies, respectively. Unfortunately, this pioneering work, formally accessible via DUA, is devalued by the fact that the 11k sentences in this corpus were arbitrarily shuffled (for increased privacy protection) so that the entire document structure has been intentionally spoiled. Hence, CARDIO:DE [56] composed of 500 clinical reports from the cardiology domain can be considered the first and only German-language clinical corpus whose structure is left intact (after de-identification) and whose accessibility is implemented via DUA as well. Since BRONCO and CARDIO:DE, follow a formalized DUA-based clearance policy they strictly adhere to internationally established distribution standards for privacy-sensitive corpora. BÖHRINGER-24 [14] composed of 300 ophthalmologic physicians' letters from three different hospitals and annotated with 2,800 diagnoses from ICD is the third in this line but raises concerns because potential clearance requires informal private negotiations which may end up in a formal DUA if permission is finally granted. On-going work on GEMTEX [59], a currently prospering major national corpus building initiative, targets an even larger (> 150k documents) and more heterogeneous collection of clinical report types covering 4 medical areas (cardiology, pathology, pharmacy, and neurology) from 6 different national clinical sites. This corpus, however, is currently not ready for use but rather stands for a corpus *in statu nascendi*. Interestingly and for the first time ever in Germany, all documents entering GeMTeX require GDPR-conformant "informed consent," i.e., the explicit agreement of patients that their clinical documents can be used (in de-identified form) for research purposes; however, potential clearance will still require some sort of DUA. These four corpora all feature standard clinical text genres (mostly discharge summaries) and are complemented by a non-standard clinical corpus, Ex4CDS [12], which is composed of 720 physicians' justifications supporting their estimated likelihood of future possible negative patient outcomes after kidney transplantations. Yet, this genre heavily drifts away from standard reporting formats we see in clinical reports and notes, and, thus, might be of minor relevance only. Thus, only 15% (5) of all document-unique real German-language clinical corpora (32) out of a total of 46 publications are open for the scientific community under most optimistic assumptions, yet only 6% (2) are currently ready for distribution under a standardized formal DUA process (comparable, e.g., with Mimic distribution standards).[6]

Once more and more single clinical/medical corpora become publicly available, potential synergies arising from their combination can be explored. Llorca *et al.* [57] describe such an approach for four corpora (Bronco, Cardio:DE, GGPOnc 2.0, and GraSCCo; the latter two will be introduced below) using the BigBIO framework [58] for (meta)data harmonization.

It is also worth noting that several attempts have been made to distribute language *models* (rather than the original non-distributable clinical raw text *data*) that were derived from classified local clinical resources (see FraMed [21], Bressem-20 [46], Roller-20 [47], Roller-22 [53], and Bressem-24 [61]). Still these detours open unexplored legal territory and face problems on their own (we will touch upon this issue below).

***Translated Real Clinical Corpora***.

Translated real clinical corpora are derived from real clinical reports and notes routinely written by professional clinical staff yet have been automatically translated from (easier to get) US-American English sources to German. We found 5 publications for such corpora from which 3 are document-unique and, also, 3 corpora are annotation-unique (actually, only 2 corpora, since one of them – N2c2-German 2.0 – differs only in terms of the number of annotated items in its most current version, not type-wise). Table 2 in the Supplementary Material section gives a detailed overview of these 5 corpora.

Becker-16 [65] relies on ShARe/CLEF eHealth 2013 Shared Task 1 resources [66] that reused Mimic-II data, whereas the n2c2-German corpus [67,68,70] builds on n2c2 2018 Shared Task Track 2 data [69] that exploited Mimic-III data. Their size is moderate (200 [65] and 400 documents [70], respectively, the latter with almost 370k tokens). Discharge summaries prevail, and the annotations relate to named entity (Disorders, Drugs) and relation extraction (Medication/Adverse Drug Events) tasks, with up to 63,4k annotation units. Idrissi-Yaghir-24 [62] make use of a much larger segment of

[6] https://physionet.org/content/mimiciii/1.4/

MIMIC-III, with 695,000k tokens after translation into German (yet without specification of the basic number of documents and without any metadata). Not as a surprise, all these corpora are publicly accessible (they inherit MIMIC's liberal DUA policy) and both versions of N2C2-GERMAN also offer a free named entity recognition model.

There are three issues with this approach. First, the quality of the automatic translation needs thorough human review by medical experts. Second, the proper alignment of the metadata must be manually validated, since begin/end positions of metadata are likely to change from English to German documents. Beyond these translation-focused issues, at a more "cultural" level, the writing style of American doctors tends to deviate from that of German ones reflecting a different reporting culture embedded in incompatible health care eco-systems. Initial experimental results on the effects of translated English documents for German clinical language models are reported by Idrissi-Yaghir *et al.* [62], although clinical data (from MIMIC-III) and non-clinical ones (from PUBMED) are indistinguishably intertwined in their experimental design.

***Synthetic Clinical Corpora***.

Synthetic clinical corpora feature invented clinical reports and notes that look like those written by professional clinical staff in terms of genre, style and terminology, but describe entirely fictitious patients and artificially constructed or massively altered medical cases. Synthetic documents are typically authored by medical experts of the same professional caliber as those authoring real ones, either by *manually writing* them from scratch or by *manually re-writing* original exemplars. With the increasing power of large language models (LLMs) rooted in the deep learning (DL) paradigm, the advent of CHATGPT [71,72] in particular, the *automatic generation* or *automatic paraphrasing* of clinical documents has become a feasible machine alternative based on prompts (instructions issued by human users which control and help tailor LLM system output). We found 6 publications for such corpora from which 3 are document-unique and 5 are annotation-unique. Table 3 in the Supplementary Material section gives a detailed overview of these 6 corpora.

JSYNCC 1.0 [73] was the first of its kind for the German clinical language and consists of 400 operative reports and 470 case reports/descriptions extracted from e-book versions of introductory textbooks for medical students. Since this corpus cannot be shared directly due to Intellectual Property Rights held by the publishers, the developers bypassed this restriction by distributing the code to reliably re-create JSYNCC copies at any other physical site (including selected metadata). As a prerequisite, the e-books incorporated in JSYNCC need to be licensed by that local institution. In the meantime, JSYNCC 2.0 [60] contains 343k annotation units covering various named entities, such as Findings, Diagnoses, Procedures, and PII.

GRASCCO can be considered a true representative of the re-writing paradigm. Despite its tiny size (63 documents, 44k tokens only), the original version, GRASCCO 1.0 [74], has developed into GRASCCO 2.0 [60] with different kinds of named entities, semantic relations, temporal relations, certainty, and negation tags, amounting to nearly 180k annotation units altogether. It is publicly accessible without any restrictions, and its most recent version, GraSCCo 3.0PHI [17] also incorporates 1,4k PII annotation units. GraSCCo is based on Austrian real discharge summaries and Web-crawled clinical documents that were massively linguistically edited, with iterative changes at the lexical, syntactic and semantic level. Furthermore, medical noise (new data items, new attribute-value sets, etc.) was intentionally injected for reasons of camouflage so that re-identification of individual patients is virtually impossible.

As to *automatic text generation* based on LLMs, Frei-23 [75] uses a prompt-based approach to generate (roughly 10k) new single sentences (*not* full-fledged documents!) which amount to slightly more than 120k tokens. An automatically generated silver standard includes 3 named entity types (Medication, Dose, and Diagnosis) comprising roughly 23k silver annotation units. As with GraSCCo, Frei-23 is publicly available without any constraints.

The motivation for and general advantage of synthetic corpora is that they circumvent the data protection problem as virtual patients and artificial cases are constructed and verbalized. Yet one may question whether synthetic documents, either written by medical experts or DL engines, sufficiently correspond with much more heterogeneous real ones and thus can really replace them without substantial analytic biases. For instance, Şerbetçi & Leser [76] report preliminary evidence that models trained on the synthetic data from Frei-23 do not transfer well to authentic clinical data from Bronco and Cardio:DE. Privacy attack experiments also revealed that reverse engineering from embeddings allows read-outs of sensitive factual data (e.g., PIIs) from LLMs via training data extraction attacks [77,78] – even in their de-identified form via a similarity search attack [79] – and therefore bear an unwanted potential for data privacy breach. Last but not least case reports, in particular those published in textbooks, deviate from authentic clinical reports in terms of a more narrative, often verbose style and non-expert language use.

### *Close Domain Proxies: Pseudo-Clinical Corpora*

*Domain proxies* for clinical corpora are collections of documents that deal with medical topics but differ from clinical reports mostly in terms of style and genre. We further refine this category in this subsection as *close* domain proxies when clinical topics are dealt with from a *scholarly* perspective at an *expert* medical level; they constitute the class of *pseudo-clinical corpora*. Perhaps the largest source of such documents is housed in PubMed-style bibliographic databases or publishers' Web portals hosting titles, abstracts or full texts of academic journal articles. Additional material comes from medical PhD theses, clinical guidelines, clinical trial reports, drug labels, or patent claims. We found 18 publications for such corpora from which 16 are document-unique and only 8 are annotation-unique. Table 4 in the Supplementary Material section gives a detailed overview of these 18 corpora.

By far the largest group composed of 13 corpora (Brown-02 [80], MuchMore [81], SpringerLink [82], Springer + MedTitle [83], Morin-12 [84], Mantra Silver [85] + Mantra GSC [86], HimL 1.0 [87], EFSG-UVigoMED [88], BTC [53], ChaDL [92], Bressem-24 [61]) makes a second-hand use of collections from bibliographic databases, such as PubMed/Medline or Livivo, or commercial publishers' websites. 8 of them are parallel/comparable multilingual corpora, with German as one of the featured languages (Brown-02, MuchMore, Springer and MedTitle, Morin-12, Mantra Silver + Mantra GSC, EFSG-UVigoMED). These proxies typically excel in huge data volumes – Mantra Silver offers the largest dataset with roughly 4,3m documents (more than 60m tokens), followed by HimL 1.0 with roughly 2,7m documents (slightly less than 60m tokens). Not surprisingly, these massive data volumes come at the price of lacking annotations. Whereas HimL 1.0 contains no metadata at all, Mantra Silver introduces the notion of a *silver standard corpus*, i.e., a huge number of automatically generated annotations as the result of harmonizing the contributions of ensembles of named entity taggers.

A second, much smaller group of corpora contains textual data from drug labels and patent claims (Mantra Silver + Mantra GSC, HimL 1.0, and ChaDL). The third one is constituted by GGPOnc which consists of clinical guidelines for oncology [90,91]. It not only stands out as a unique guideline corpus publicly available via DUA, but is large-sized (about 10k text segments from the complete set of 30 German oncology guidelines, with roughly 1,900k tokens) and excels in annotations with either 7 named entity types (GGPOnc 1.0 [90]) taken from the UMLS Semantic Groups (with around 73,8k annotation units) or 3 Snomed CT-anchored named entity types (GGPOnc 2.0 [91]), currently summing up to roughly 450k curated annotation units [91,60].

Scholarly writing is fundamentally different from clinical writing – not only in terms of genre and style, but also in terms of language use characteristics. Whereas scholarly articles mostly adhere to linguistic well-formedness, terminological canonicity and definitional clarity, clinical reports abound with paragrammatical syntax, spelling errors, local clinical jargon (exemplified by in-house abbreviations or acronyms) typical of language performance under high work load and, thus, heavy time pressure, as well as closed language community conventions. Perhaps the main difference, however, lies in their diverging communicative intention – whereas academic writing usually addresses the generalizability of observables (e.g., the effect of a drug or a medical procedure within a patient cohort), clinical reports focus on individual patients only. Whether these considerations have a measurable impact on training or adapting language models remains an issue of further investigations.

### ***Distant Domain Proxies: Non-Clinical Medical Corpora***

*Distant* domain proxies for clinical corpora are sets of documents covering medical topics from a non-clinical perspective, targeting mainly non-expert comprehensibility, here referred to as *non-clinical medical corpora*. In this group, the genre-specific style of clinical reporting vanishes completely, although lexical adherence to medical terminology is sought for, often at a layman level (e.g., “*Blinddarmentzündung*” is preferred over “*Appendicitis*”, “*Blutvergiftung*” over “*Sepsis*”). We found 17 publications for such corpora from which all 17 are document-unique whereas 13 are annotation-unique. Table 5 in the Supplementary Material section gives a detailed overview of these 17 corpora.

The dominant group of distant domain corpora is composed of 10 resources in which *social media* data are assembled, either incorporating medically focused chats extracted from general social media platforms, such as Twitter or Telegram [96,97,100], or from thematically specialized public health portals, e.g., dealing with diabetes, obesity, drug misuse, or depression. Though a layman language attitude prevails in this *dialogical* data, medical expert statements can be found here as well, particularly in public health portals, yet rigorous medical expert jargon is typically avoided. Exemplars of social media medical corpora are TLC-MED1 [94] which collects excerpts from the German MED1.DE health portal, BECK-21 [96] in which Covid-19-related messages were collected, LIFELINE [98,99] which contains threads thematically related to adverse drug reactions, BTC [53], BRESSEM-24 [61], HEINRICH-24 [100] (compiling conspiracy narratives within the Covid discourse), and HEALTHFC [101] (a claim–evidence–verdict triple dataset for fact checking). The data volume varies a lot in this category – from few hundred thousand tokens (TLC-MED1) via half a million for LIFELINE, up to more than 9m tokens in BRESSEM-24.

A second class of distant domain corpora is formed by 5 resources composed of (*monological*) online *encyclopedic articles* as available, e.g., from WIKIPEDIA. Typical examples of this approach are, e.g., WIKISECTION [93] (basically a disease corpus), CHADL [92], BRESSEM-24 [61], or FREI-24 [103]. These are also high-volume datasets, with 2k-4k documents (2m-3m tokens), CHADL with more than 20m tokens being the largest one.

Finally, perhaps the most distant, collections of general *newspaper/newswire articles* are assembled in corpora dealing with medical themes, such as LOHR-16 [31], RSS [95], and FANG-COVID [97]. These corpora are typically supersized, with (tens up to hundreds of) millions of tokens, yet without any metadata.

Not surprisingly, all these corpora are publicly available although care should be taken when social media data are chosen, e.g., from health consultation or disease community portals, where privacy issues easily pop up [104,105]. Distant domain proxies are typically large-sized, with millions of tokens, yet often lack deeper medical metadata (WIKISECTION, TLC, BECK-21, LIFELINE 2.0, HEINRICH-24, HEALTHFC, and FREI-24 being notable exceptions from this rule). Fundamental concerns may be raised whether these sources can reasonably be used, at all, as a substitute for clinical data due to heavily divergent genre, style, argumentation, and vocabulary patterns.

## DISCUSSION

Corpora are an indispensable prerequisite for training, tuning, adapting, and evaluating (large) language models.[7] In the clinical domain, however, these resources are hard to get because of ethical concerns that have been translated into rigorous data protection laws world-wide. In Germany, for instance, at the time of this writing (January 2025) 27 non-distributable, yet often richly annotated clinical datasets are kept in closed local data silos inaccessible for clinic-external researchers. This constitutes not only an enormous waste of money and human resources, but also a serious loss of medical opportunities for better diagnosis, treatment, quality of life and, last but not least, an increase of survival chances of hospitalized patients (not to mention the reduction of costs for the health care system). Fortunately, this (over-)protective siloing strategy is starting to become more permeable as witnessed by the strictly *DUA*-formalized accessibility of the Cardio:DE [56] and Bronco [11] clinical report corpora. Three additional corpora may be counted as potential alternatives – Böhringer-24 [14] (though the access option rests on a fluffy and only informal distribution offer), GeMTeX [59] (a corpus building initiative just launched, the results of which will only be available during 2025, but are fully compliant with EU regulations (GDPR) based on informed consent), and Ex4CDS [12] whose domain of discourse (risk justifications after kidney transplantations) is somewhat off topic compared with standard clinical reports and notes.

[7] Activities related to generating clinical German-language models that make use of many of the corpora introduced in this review are reported, e.g., in [92,106,107,10,108,60,14,13,109].

Several researchers offer a substitute in that they do not distribute locked clinical *raw data* or associated *metadata* but rather allow the *language models* generated from these original data to be distributed. One caveat must be made – data privacy issues may pop up here since evidence has been reported that individual patients' data can indeed be read out from the models' representation structures and thus bear the danger of patient re-identification demanding further safety measures against hostile attacks [77-79].

That said, we also looked at alternative corpus designs that have been investigated to escape from clinical data sparsity. We organized these efforts in a taxonomy based on qualitative considerations. For *real* clinical corpora, we found two ways to circumvent data access restrictions. The first one is to pick up DUA-accessible English data and *translate* them automatically. The second strategy is to generate, manually or automatically, *synthetic* clinical reports with fictitious content.

As another alternative, we identified *domain proxies* for clinical reports. They deal with clinical or, more general, medical, topics written by medical experts or laymen, yet depart from standard clinical report writing in terms of genre, style and terminology to a varying degree though. The category of *close* domain proxies is constituted by pseudo-clinical documents, such as the whole range of scientific medical literature (abstracts and full texts from journals), therapy guidelines, clinical trial reports, drug labels/leaflets or patent claims. Yet, also more *distant* domain proxies play a role here, namely those that deal with medical themes without clinical phrasing, because their target is a general, non-expert audience. This category is filled by chats, threads or tweets from generic social media channels or specialized health portals, or by encyclopedic articles from Wikipedia. Altogether (see Table 6), we identified 71 distinct, i.e., document-unique, and 69 annotation-unique German-language corpora from 92 publications.[8]

| **Corpus Type** | **Different corpus versions** | **Document-unique corpora** | **Annotation-unique corpora** |
|---|---|---|---|
| **Clinical – real** | FraMed [21] | FraMed [21] | FraMed [21] |
| | Müller-07 [24] | Müller-07 [24] | – |
| | Spat-08 [25] | Spat-08 [25] | Spat-08 [25] |
| | Kreuzthaler-11 [26] | Kreuzthaler-11 [26] | Kreuzthaler-11 [26] |
| | Fette-12 [27] | Fette-12 [27] | Fette-12 [27] |
| | Bretschneider-14 [29] | Bretschneider-14 [29] | Bretschneider-14 [29] |
| | | *SuperSet-of* | |
| | Bretschneider-13 [28] | Bretschneider-13 [28] | Bretschneider-13 [28] |

[8] Some corpora were assigned to more than one of the five categories. Therefore, this publication count (92) is higher than the number of relevant hits (78).

| | | | |
|---|---|---|---|
| | Toepfer-15 [30] | Toepfer-15 [30] | Toepfer-15 [30] |
| | Lohr-16 [31] | Lohr-16 [31] | Lohr-16 [31] |
| | Löpprich-16 [32] | Löpprich-16 [32] | Löpprich-16 [32] |
| | Roller-16 [33] | Roller-16 [33] | Roller-16 [33] |
| | Roller-20 [47] | = Roller-20 [47] | Roller-20 [47] |
| | | *SuperSet-of* | |
| | Roller-22 [53] | Roller-22 [53] | Roller-22 [53] |
| | Cotik-16 [35] | Cotik-16 [35] | Cotik-16 [35] |
| | Roller-18 [37] | Roller-18 [37] | Roller-18 [37] |
| | Kreuztaler-16 [34] | Kreuztaler-16 [34] | Kreuztaler-16 [34] |
| | Seuss-17 [16] | Seuss-17 [16] | Seuss-17 [16] |
| | Oleynik-17 [36] | Oleynik-17 [36] | – |
| | Krebs-17 [38] | Krebs-17 [38] | Krebs-17 [38] |
| | 3000PA 5.0 [60] | 3000PA 5.0 [60] | 3000PA 5.0 [60] |
| | | *SuperSet-of* | |
| | 3000PA 1.0 [39] | 3000PA 1.0 [39] | 3000PA 1.0 [39] |
| | | *SuperSet-of* | |
| | 3000PA 2.0 [40] | 3000PA 2.0 [40] | 3000PA 2.0 [40] |
| | 3000PA 3.0 [42] | 3000PA 3.0 [42] | 3000PA 3.0 [42] |
| | 3000PA 4.0 [45] | 3000PA 4.0 [45] | 3000PA 4.0 [45] |
| | Becker-19 [41] | Becker-19 [41] | Becker-19 [41] |
| | Cardio:DE [56] | Cardio:DE [56] | Cardio:DE [56] |
| | | *SuperSet-of* | |
| | CardioAnno [49] | CardioAnno [49] | CardioAnno [49] |
| | | *SuperSet-of* | |
| | Richter-Pechanski-19 [43] | Richter- Pechanski-19 [43] | Richter-Pechanski-19 [43] |
| | König-19 [44] | König-19 [44] | König-19 [44] |
| | Bressem-24 [61] | Bressem-24 [61] | Bressem-24 [61] |
| | | *SuperSet-of* | |
| | Bressem-20 [46] | Bressem-20 [46] | Bressem-20 [46] |
| | Grundel-21 [48] | Grundel-21 [48] | Grundel-21 [48] |
| | Bronco [11] | Bronco [11] | Bronco [11] |
| | MedCorpInn [50] | MedCorpInn [50] | – |
| | | *SuperSet-of* | |
| | MedCorpInn$_{SUB}$ [50] | MedCorpInn$_{SUB}$ [50] | – |
| | | *SuperSet-of* | |
| | Karbun [51] | Karbun [51] | – |
| | Madan-22 [52] | Madan-22 [52] | Madan-22 [52] |
| | [Ex4CDS] [12] | [Ex4CDS] [12] | [Ex4CDS] [12] |
| | Trienes-22 [55] | Trienes-22 [55] | Trienes-22 [55] |
| | Llorca-23 [57] | Llorca-23 [57] | Llorca-23 [57] |
| | GeMTeX [59] | GeMTeX [59] | GeMTeX [59] |
| | Böhringer-24 [14] | Böhringer-24 [14] | Böhringer-24 [14] |
| | Idrissi-Yaghir-24 [62] | Idrissi-Yaghir-24 [62] | – |
| | RadQA [62] | RadQA [62] | RadQA [62] |
| | DMP “HerzMobil” [63] | DMP “HerzMobil” [63] | DMP “HerzMobil” [63] |
| | Plagwitz-24 [64] | Plagwitz-24 [64] | Plagwitz-24 [64] |
| | Σ: 46 | Σ: 32 | Σ: 40 |
| **Clinical – translated** | Becker-16 [65] | Becker-16 [65] | Becker-16 [65] |
| | N2c2-German 2.0 [70] | N2c2-German 2.0 [70] | N2c2-German 2.0 [70] |
| | | *SuperSet-of* | |
| | N2c2-German 1.0 [67,68] | N2c2-German 1.0 [67,68] | N2c2-German 1.0 [67,68] |
| | Idrissi-Yaghir-24 [62] | Idrissi-Yaghir-24 [62] | – |
| | Σ: 5 | Σ: 3 | Σ: 3 |
| **Clinical – synthetic** | JSynCC 2.0 [60] | JSynCC 2.0 [60] | JSynCC 2.0 [60] |
| | | *SuperSet-of* | |
| | JSynCC 1.0 [73] | JSynCC 1.0 [73] | JSynCC 1.0 [73] |
| | GraSCCo 1.0 [74] | GraSCCo 1.0 [74] | – |
| | GraSCCo 2.0 [60] | = GraSCCo 2.0 [60] | GraSCCo 2.0 [60] |
| | GraSCCo 3.0$_{PHI}$ [17] | = GraSCCo 3.0$_{PHI}$ [17] | GraSCCo 3.0$_{PHI}$ [17] |
| | Frei-23 [75] | Frei-23 [75] | Frei-23 [75] |
| | Σ: 6 | Σ: 3 | Σ: 5 |
| **Close domain proxies** | Brown-02 [80]<br>MuchMore [81]<br>Springer-Link [82]<br>Springer [83]<br>MedTitle [83]<br>FraMed [21]<br>Morin-12 [84]<br>Mantra [Silver] [85]<br><br>Mantra GSC [86]<br>HimL 1.0 [87]<br>EFSG-UVigoMED [88]<br>Villena-20 [89]<br>GGPOnc 2.0 [91]<br><br>GGPOnc 1.0 [90]<br>BTC [53]<br>ChaDL [92]<br>Bressem-24 [61]<br>Idrissi-Yaghir [62]<br>Σ: 18 | Brown-02 [80]<br>MuchMore [81]<br>Springer-Link [82]<br>Springer [83]<br>MedTitle [83]<br>FraMed [21]<br>Morin-12 [84]<br>Mantra [Silver] [85]<br>*SuperSet-of*<br>Mantra GSC [86]<br>HimL 1.0 [87]<br>EFSG-UVigoMED [88]<br>Villena-20 [89]<br>GGPOnc 2.0 [91]<br>*SuperSet-of*<br>GGPOnc 1.0 [90]<br>BTC [53]<br>ChaDL [92]<br>Bressem-24 [61]<br>Idrissi-Yaghir [62]<br>Σ: 16 | –<br>MuchMore [81]<br>Springer-Link [82]<br>–<br>–<br>FraMed [21]<br>–<br>Mantra [Silver] [85]<br><br>Mantra GSC [86]<br>–<br>EFSG-UVigoMED [88]<br>–<br>GGPOnc 2.0 [91]<br><br>GGPOnc 1.0 [90]<br>–<br>–<br>–<br>–<br>Σ: 8 |
| **Distant domain proxies** | FraMed [21]<br>Lohr-16 [31]<br>ML–UVigoMED [88]<br>WikiSection [93]<br>TLC-Med1 [94]<br>RSS [95]<br>Beck-21[96]<br>Fang-Covid [97]<br>Lifeline 1.0 [98]<br>BTC [53]<br>ChaDL [92]<br>Bressem-24 [61]<br>Lifeline 2.0 [99]<br>Heinrich-24 [100]<br>HealthFC [101]<br>Pedrini-24 [102]<br>Frei-24 [103]<br>Σ: 17 | FraMed [21]<br>Lohr-16 [31]<br>ML–UVigoMED [88]<br>WikiSection [93]<br>TLC-Med1 [94]<br>RSS [95]<br>Beck-21[96]<br>Fang-Covid [97]<br>Lifeline 1.0 [98]<br>BTC [53]<br>ChaDL [92]<br>Bressem-24 [61]<br>Lifeline 2.0 [99]<br>Heinrich-24 [100]<br>HealthFC [101]<br>Pedrini-24 [102]<br>Frei-24 [103]<br>Σ: 17 | FraMed [21]<br>Lohr-16 [31]<br>ML–UVigoMED [88]<br>WikiSection [93]<br>TLC-Med1 [94]<br>RSS [95]<br>Beck-21[96]<br>Fang-Covid [97]<br>Lifeline 1.0 [98]<br>–<br>–<br>–<br>Lifeline 2.0 [99]<br>Heinrich-24 [100]<br>HealthFC [101]<br>–<br>Frei-24 [103]<br>Σ: 13 |
| **Overall** | Σ: 92 | Σ: 71 | Σ: 69 |

**Table 6:** Summary of German-Language Clinical/Medical Corpora

Idrissi-Yaghir–24 [62] is currently by far the largest of all German-language medical corpora, with slightly more than 25m documents and 3,0b tokens from its clinical segment, plus the translated Mimic-III clinical segment (695,000k tokens), plus the translated PubMed segment (6,000k abstracts with 1,700,000k tokens) – roundabout more than 31m documents with 5,4b tokens. The vast clinical portion of this corpus is used for in-house training of the language model – a recent trend leading to in-house, i.e., hospital-specific, language models without the need for de-identification and data sharing. Bressem-24 [61] is even more heterogeneous and the second-largest medical German-language corpus, a hybrid conglomerate of clinical reports, embedded public corpora (GGPOnc, GraSCCo), a PubMed subset, publisher-provided scientific papers, and medical PhD theses – overall, more than 4,7m documents (1,1b tokens).

Their sheer amount of tokens is truly impressive, yet when it comes to the supply of clinically relevant metadata, other corpora deserve equal credit. On this dimension, we find

- 3000PA 5.0 [60], with 6,600 documents (7,300k tokens) and 2,093k multi-level annotation units, including section, named entity and relation annotations, as well as annotations involving temporality and factuality,
- Cardio:DE [56], with 500 documents (993k tokens) and 143,5k named entity and relation annotations,
- Roller-20 [47], with 1,725 documents (158k tokens) with 77,4k named entity and relation annotations.

Among these three corpora, Cardio:DE stands out as the only one that is accessible on a formalized DUA basis (together with the smaller and less richly annotated Bronco corpus).

Still the taxonomy we introduced leaves an important issue open: How close/distant, in a metrical sense, are potential substitutes when compared with real clinical reports in terms of genre, style, jargon and diction? This *stylometric* question should be complemented by a *functional* one: How good are these substitutes in terms of classification performance when compared to real clinical documents? Initial attempts at answering this emerging research question have already been made. Modersohn *et al.* [74] compared a synthetic clinical corpus (GraSCCo) with a real one (3000PA) by clustering syntactic and semantic features, whereas Lohr & Hahn [110] developed DoPA Meter, a stylometric toolkit with more than 120 style metrics covering lexical, syntactic and semantic expression layers, and ran it on synthetic, as well as on close and distant domain proxies. However, a comprehensive functional comparison is still lacking although first experiments have been reported for Cardio:DE, Bronco, GGPOnc 2.0, and GraSCCo by Llorca *et al*. [57] and Şerbetçi & Leser [76]. Stylometric analyses could highlight descriptive differences in terms of linguistic variance whereas an experimental comparison of the (classification) performance of language models trained on real clinical corpora with ones trained on translated, synthetic and proximal substitutes could lead to an empirically founded “cost model” for corpus substitution.

**Acknowledgments.**

First, and foremost, I want to thank the reviewers for their detailed and extremely helpful comments and suggestions. The revised version of the original submission reflects their proposals in many ways. Second, my thanks go to Christina Lohr who commented on the draft version in a very helpful way. Finally, Frank Meineke provided me with details about the enormous variety of text genres in the clinical domain from a real-life perspective.

**Competing Interests.**

The author declares that there are no competing interests.

**Funding.**

The author was and is currently funded by the German *Bundesministerium für Bildung und Wissenschaft (BMBF)* under grants SMITH (01ZZ1803G) and GeMTeX (01ZZ2314B), respectively.

**SUPPLEMENTARY MATERIAL**

Supplementary material is available at JAMIA Open online.

**CONFLICT OF INTEREST STATEMENT**

None.

**DATA AVAILABILITY**

# Supplementary Material

## A. Tables of German-Language Clinical/Medical Corpora

The five tables contained in this section are grouped, in decreasing order, by typological homogeneity and text genre similarity relative to German-language clinical/medical reports and notes:

- **Table 1** features original *clinical* corpora composed of authentic textual material (e.g., discharge summaries, pathology or radiology reports, etc.),
- **Table 2** lists clinical corpora that have been *translated* from a foreign language (typically, American English) to German,
- **Table 3** introduces *synthetic* clinical corpora with fictitious descriptions of virtual patients, yet in the format of original clinical documents,
- **Table 4** assembles non-clinical corpora with medical contents though, collected from *scholarly publications* hosted in digital libraries (typically, PubMed), publishers' web sites, or even science-focused newspaper articles,
- **Table 5** contains non-clinical corpora that were built from *encyclopedic articles* (typically, Wikipedia) or *social media* data (tweets, chats, blogs, etc.), all dealing with medical topics.

Each of these tables is structured following a common column format:

- The first column contains the *name* of the corpus (if explicitly introduced in the cited publications) or a pseudo name (first author plus publication year), the *citation*, and the *year of publication* (the rows are ordered in ascending order by year),
- The second column specifies the *number of documents* in the corpus,
- The third column indicates the *number of tokens* in the corpus,
- The fourth column lists the *document type(s) or text genre(s)* incorporated in the corpus, including the specific medical domain the documents deal with,
- The fifth column provides detailed information of the *metadata* that was added to the documents, i.e., the annotation types and number of associated annotation items provided; in addition, we indicate whether
    - *entity normalization*, i.e., grounding of the entity instances in some (medical) terminology or ontology, was carried out (and, if so, mention the chosen concept system,
    - *annotation guidelines* that were used for the manual generation of metadata are accessible (e.g., in the supplementary material section of the article, or in and open access data portal, such as GitHub),
    - *inter-annotator agreement (IAA)* was measured by some canonical metric and the resulting scores are reported as individual values per type or in an aggregated macro form over all types,
- The sixth column marks the availability of the corpus, i.e., whether it is *inaccessible* (classified): 🔴, *publicly available via contract-based access*, typically based on a Data

Use Agreement (DUA), or other types of private commitments or institutional negotiations: ✉ ), or *publicly available without any restrictions*: ✓ ; as a special option, ◆ marks the availability of language models derived from a specific corpus.[9]

"noi" indicates that "no information" about specific quantitative data is reported in the publication; "n/a" indicates that a specific categorical information is "not applicable" (e.g., IAA data for automatically generated (silver standard) metadata or data extracted from the structured portion of the EHR segment of the clinical information system.

Descriptions that are relevant for a particular category (say, real clinical reports) are colored in black whereas grey colored areas apply to other categories of the same corpus (e.g., if the corpus contains Wikipedia data, as well, which are discussed in the table relating to distant domain proxies). Accordingly, the reader always gets a complete picture of the textual variety of the corpus without loosing focus.

In **Appendix A** we propose a more elaborate corpus datasheet, the template for a corpus card, with mandatory and (desirable) optional description categories for clinical/medical corpora (see Table 7).

[9] The distinctions from above deserve some further clarifications. Corpora were classified as "inaccessible" if either accessibility was explicitly denied in the publication, or public distributability was not all mentioned and no de-identification efforts were reported. "Public availability" has two decision branches. The first one either relies on a *formal* procedure, mostly based on contractual DUAs, or on *informal* (private) commitments which leave open (but at the same time do not explicitly preclude) whether access will be granted. Admittedly but intentionally, this is a soft constraint for distribution permissions. The second branch, "public availability without restrictions" simply holds if the corpus comes with a valid physical address for download from a digital host (e.g., institutional directories, open resource distribution sites such as GitHub or Zenodo, etc.).

| CORPUS / Citation – Year | Documents | Tokens (in 1,000 =1k) | Clinical Document Types (Text Genres) | Metadata | Availability ● Corpus ◆ Model |
|---|---|---|---|---|---|
| FRAMED [21] – 2004 | noi (~6,500 sentences) | 100k | Various clinical report types (discharge, pathology, histology, and surgery reports), a medical textbook, and Web documents taken from a consumer health care portal (netdoktor) | **Annotation Types**<br>Sentence & token splits, parts of speech (PoS)<br>**Entity Normalization:** Y (medically adapted STTS for PoS annotation)<br>**Annotation Guideline:** N<br>**IAA Measurement:** Y | ● ◆ FRAMED model as part of JCORE[10] [22,23] |
| MÜLLER-07 [24] – 2007 | ~ 30,000 | noi | Mainly discharge letters, but also surgical reports, immunodermatological findings and other narrative reports of clinical results (dermatology) | none | ● |
| SPAT-08 [25] – 2008 | 1,500<br>subset from 18k | noi | 26 clinical document types from 8 medical fields (vascular & casualty surgery, internal medicine, neurology, anaesthesia, intensive care, radiology, physiotherapy) | **Annotation Types**<br>Classification into document types and medical fields<br>**Entity Normalization:** N<br>**Annotation Guideline:** N<br>**IAA Measurement:** N | ● |
| KREUZTHALER-11 [26] – 2011 | 3,542 | 84k | Pathology reports | **Annotation**<br>(Automatic) rewriting of fully capitalized texts as mixed capitalized and lower-cased texts (following German orthography rules)<br>**Entity Normalization:** n/a<br>**Annotation Guideline:** n/a<br>**IAA Measurement:** n/a | ● |
| FETTE-12 [27] – 2012 | 544<br>subset from 193k | noi | Clinical reports from 5 clinical domains (echocardiography, ECG, lung function, X-ray thorax, bicycle stress test) | **Annotation Types**<br>Automatic extraction of attribute-value pairs from the 5 clinical domains<br>**Entity Normalization:** Y (local terminology)<br>**Annotation Guideline:** N<br>**IAA Measurement:** N | ● |
| BRETSCHNEIDER-13 [28] – 2013 | 174<br>subset from 2,7k | 28k | Radiology reports (lymphoma) | **Annotation**<br>classification into "*pathological*" or "*non-pathological*" sentences<br>**Entity Normalization:** N<br>**Annotation Guideline:** N<br>**IAA Measurement:** N | ● |
| BRETSCHNEIDER-14 [29] – 2014 | 2,713 | 347k | Radiology reports (lymphoma) | **Annotation Types**<br>**(Annotated items)**<br>Automatic concept annotation with (machine-translated) German RADLEX terms ($\Sigma$: 148k tokens (= 42.6 %))<br>**Entity Normalization:** Y (RadLex)<br>**Annotation Guideline:** n/a<br>**IAA Measurement:** n/a | ● |

[10] https://julielab.de/Resources/JCoRe.html

| | | | | | |
|---|---|---|---|---|---|
| Toepfer-15<br>[30] – 2015 | 140<br>subset from 69k/70k | noi | (Transthoracic) echocardiography reports | **Annotation Types**<br>**(Annotated items)**<br>Automatic extraction of 440 attribute-value pairs from the echocardiography domain (e.g., attributes: *Aortic Valve, Mitral Valve, Tricuspid Valve, regurgitation (aortic), Aorta, Stenosis, Diastolic Function, Left/Right Ventricle*; values: *present, absent, severe*)<br>(Σ: 6,2k)<br>**Entity Normalization:** Y (local terminology mapped to a guideline for German transthoracic echocardiography reports)<br>**Annotation Guideline:** N<br>**IAA Measurement:** Y | ● |
| Lohr-16<br>[31] – 2016 | 450<br>subset from 22,4k<br><br>5,8m | 266k<br><br>125,9m | Operative reports (digestive tract)<br><br>(Fragments of) newspaper articles with medical content extracted from DWDS (*Digitales Wörterbuch der Deutschen Sprache*) | **Annotation Types**<br>*Diagnoses*, *Procedures*<br>**Entity Normalization:** Y (ICD for diagnoses, OPS for executed procedures)<br>**Annotation Guideline:** n/a (extracted from EPR as gold standard)<br>**IAA Measurement:** n/a<br><br>Mentions of 400 medical terms, such as *“patient”, “surgery”, “ambulance”*, etc.<br>**Entity Normalization:** N<br>**Annotation Guideline:** n/a<br>**IAA Measurement:** n/a | ● |
| Löpprich-16<br>[32] – 2016 | 737<br>(paragraphs only) | noi | Main diagnosis paragraphs split from discharge summaries (*oncology*: *multiple myeloma*) | **Annotation Types**<br>**(Annotated items)**<br>*Diagnosis* (0,9k), *State of Disease* (specific data elements characteristic for multiple myeloma; 7,7k)<br>(Σ: 8,6k)<br>**Entity Normalization:** N<br>**Annotation Guideline:** N<br>**IAA Measurement:** Y | (✔ \| )<br>● [11] |
| Roller-16<br>[33] – 2016 | 118<br>+ 1,607<br>= 1,725 | 90k<br>+ 68k<br>= 158k | Discharge summaries & clinical notes (nephrology) | **Annotation Types**<br>**(Annotated items)**<br>23 entity types, grouped into 7 major categories:<br>[Time: *Date, Temporal Course*;<br>Person/Body: *Person, Body Part, Tissue, Body Fluid, Localization*;<br>Process: *Process*;<br>Condition: *State of Health, Medical Condition, Diagnostic/Lab Procedure, Medical Specification, Degree, Type*;<br>Therapy: *Medical Device, Medication, Biological Chemistry, Treatment, Measurement*;<br>Structure: *Structure Element*;<br>Factuality: *Modality Positive, Modality Negation, Modality Vagueness*] | ● |

[11] The corpus has been announced to be publicly available in the supplementary online files of the publication. However, upon inspection of the supplement, the corpus was not listed. Further email communication with the authors revealed that this statement was way too optimistic. In conclusion, the corpus cannot be distributed.

| | | | | | |
|---|---|---|---|---|---|
| | | | | **Entity Normalization:** Y (UMLS)<br>**Annotation Guideline:** N (scheme only)<br>**IAA Measurement:** N | |
| Kreuztaler-16<br>[34] – 2016 | 1,696 | noi | Discharge letters (dermatology) | **Annotation**<br>**(Annotated items)**<br>abbreviated word forms<br>(Σ: 2,3 k)<br>**Entity Normalization:** N<br>**Annotation Guideline:** N<br>**IAA Measurement:** Y | ● |
| Cotik-16<br>[35] – 2016 | 8<br>+ 175<br>= 183 | 6,2k<br>+ 6,7k<br>= 12,9k | Discharge summaries & clinical notes (nephrology) | **Annotation Types**<br>**(Annotated items)**<br>*Negation* (0,4k) & *Factuality*: *affirmed* (0,6k), *speculated* (<0,1k) of *Findings*<br>(Σ: 1,1k)<br>**Entity Normalization:** Y (UMLS)<br>**Annotation Guideline:** N (schema only)<br>**IAA Measurement:** N | ● |
| Seuss-17<br>[16] – 2017 | 1,400<br>[subset from 4,671<br>+ 2,804<br>+ 1,008<br>+ 6,223<br>= 14,706 | ~5,000k<br>~50,000k | pathology reports<br>medical reports (Gynecology)<br>operative reports (Gynecology)<br>radiology reports | **Annotation Types**<br>**(Annotated items)**<br>9 Personally Identifiable Information (PII) types<br>[*Name, Age, Contact, Address, Date of birth/surgery/ examination, Medical ID,* etc.]<br>(Σ: 23,5k)<br>**Entity Normalization:** N<br>**Annotation Guideline:** N<br>**IAA Measurement:** N | ● |
| Oleynik-17<br>[36] – 2017 | 30,000 | noi | discharge summaries (cardiology) | none<br>(200 abbreviations) | ● |
| Roller-18<br>[37] – 2018 | 626<br>(subset from [33]) | 26,5k*<br>(*estimated from averages) | Clinical notes & discharge summaries (nephrology) | **Annotation Types**<br>**(Annotated items)**<br>8 named entity types:<br>[*Medical Condition: Symptom*, *Finding*, *Diagnosis* (2,5k), *Treatment* (1,7k), *State of Health* (1,5k), *Medication* (1,2k), *Biological Process* (1,2k), *Body Part / Organ* (0,8k), *Medical Specification* (0,8k), Locality (1,9k)]<br>(Σ: 9,7k)<br>5 relation types:<br>[*hasState* (0,4k), *Involves* (0,4k), *hasMeasure* (0,4k), *isLocated* (0,2k), *isSpecified* (0,1k)]:<br>(Σ: 1,5k)<br>**Entity Normalization:** N<br>**Annotation Guideline:** N (scheme only)<br>**IAA Measurement:** N | ● |
| Krebs-17<br>[38] – 2017 | | | Radiology reports (chest) | 1. Semi-automatic acquisition of a local clinical terminology composed of 258 attributes for processing radiology reports;<br>2. Value categories for attributes: *negation*, *laterality* (right, left, both sides), *location*, *degree of severity*, *condition-after*, & *progression note*. | ● |
| | 100<br>subset from 3,000 | noi | | 3. Automatic extraction of 735 attribute-value pairs.<br>**Entity Normalization:** Y (local terminology)<br>**Annotation Guideline:** n/a<br>**IAA Measurement:** n/a | |
| 3000PA 1.0<br>[39] – 2018 | 2,360<br><br>(from 3 different clinical sites) | 3,997k | (mostly) Discharge summaries, few transfer letters | **Annotation Types**<br>1 *Medication* entity + 5 *Medication* relation types<br>[*Medication/Drug: Dosage, Mode, Frequency, Duration, Medical Reason*]<br>**Entity Normalization:** N<br>**Annotation Guideline:** N (scheme only)<br>**IAA Measurement:** Y | ● |
| 3000PA 2.0<br>(1000PA-J)<br>[40] – 2018 | 1,106<br><br>subset from 3000PA | 1,500k | (mostly) Discharge summaries, few transfer letters | **Annotation Types**<br>**(Annotated items)**<br>18 *Section Heading* types<br>[*Salutation* (12,9k), *Anamnesis* (0,6k): *Patient history (6,0k)* & *Family history* (<0,1k), *Diagnosis* (4,0k): *Admission diagnosis* (9,2k) & *Discharge diagnosis* (4,8k), *Hospital discharge studies summary* (87,1k), *Procedures* (3,9k), *Allergies intolerances risks* (0,2k), *Medication* (0,4k): *Admission medication* (0,1k) & *Medication during stay* (0,5k) & *Discharge medication* 11,6k), *Hospital course* (19,8k), *Plan of care* (3,6k), *Final remarks* (4,8k), *Supplements* (1,0k)]<br>(Σ: 171k)<br>**Entity Normalization:** Y (CDA-compliant)<br>**Annotation Guideline:** N (scheme only)<br>**IAA Measurement:** Y | ● |
| BECKER-19<br>[41] – 2019 | 820<br>+ 817<br>+ 107<br>+ 326<br>+ 20<br>+ 423<br>= 2,513<br>subset from 5,506 | noi | (Mixed) clinical reports:<br>medical reports,<br>radiology reports,<br>microbiology reports,<br>pathology reports,<br>virology reports, and<br>tumor board protocols | **Annotation Types**<br>**(Annotated items)**<br>11 named entity types, attributes and values related to *colorectal cancer*<br>[*ICD-Code* (0,4k), *TNM staging (0,6k), distance measurements (0,1k), microsatellite instability (0,1k), resection potential (0,3k), mutation status* (0,2k), *intensive therapy* (< 0,1k), *large tumor burden* (< 0,1k), *rapid progress* (< 0,1k), *tumor symptoms* (< 0,1k), *organ complications* (0,2k]<br>(Σ: 2,0k)<br>**Entity Normalization:** Y (UMLS)<br>**Annotation Guideline:** N (scheme only)<br>**IAA Measurement:** Y | ● |
| 3000PA 3.0<br>(1000PA-J)<br>[42] – 2019 | 1,106<br>subset from 3000PA | 1,400k | (mostly) Discharge summaries, few transfer letters | **Annotation Types**<br>**(Annotated items)**<br>13 *PII* types<br>[*Age* (0,5k), *Contact (phone, email, URL;* 0,6k*), Date* (20,6k), *Birthdate (1,1k), ID (patient, e.g., EPR number;* 0,4k*; Typist*; 0,7k*), Location (physical address*; 5,4k*), Medical Unit (hospital or department name*; 6,2k*), Person* (<0,1k), *Patient* (3,2k), *Relative* (<0,1k), *Staff* (5,2k), *Other* (0,2k)] (Σ: 44,2k)<br>**Entity Normalization:** N<br>**Annotation Guideline:** N (scheme only)<br>**IAA Measurement:** Y | ● |
| RICHTER-PECHANSKI-19<br>[43] – 2019 | 113 | 107k | Medical reports (cardiology) | **Annotation Types**<br>**(Annotated items)**<br>8 *PII* types<br>[*person, location, date, phone, organization, title, salutation, zip code*]:<br>(Σ: 5,2k)<br>**Entity Normalization:** N<br>**Annotation Guideline:** N<br>**IAA Measurement:** N | ● |
| KÖNIG-19<br>[44] – 2019 | 1,982 | 2,001k | Discharge summaries (osteoporosis) | **Annotation Types**<br>**(Annotated items)**<br>1 *Drug-Disease* relation<br>[“*proton-pump inhibitor use – osteoporosis*”] (2,0k), including concept recognition for *PPI* and *osteoporosis* [extracted from the hospital-internal study database as gold standard]<br>**Entity Normalization:** Y (Wingert Nomenclature)<br>**Annotation Guideline:** n/a<br>**IAA Measurement:** n/a | ● |
| 3000PA 4.0 (1000PA-J)<br>[45] – 2020 | 1,106<br><br>subset from 3000PA | 1,500k | (mostly) Discharge summaries, few transfer letters | **Annotation Types**<br>**(Annotated items)**<br>3 named entity types<br>[*Diagnosis* (55k), *Findings* (155k), *Symptoms* (8k)] &<br>3 attributes of NE types<br>[*Time* (previous, recurrent, uncertain), *Modality* (suspected, excluded, uncertain), *Complexity*]<br>**Entity Normalization:** N<br>**Annotation Guideline:** N<br>**IAA Measurement:** Y | ● |
| BRESSEM-20<br>[46] – 2020 | 5,783<br><br>subset from 3,8m radiology reports used for model pre-training | 399k*<br>(*estimated from averages)<br><br>416m | Radiology reports (chest radiographs, chest CT scans) | **Annotation Types**<br>**(Annotated items)**<br>9 *Finding* types, incl. *Medical Devices*<br>[*Congestion* (1,5k), *Opacity (e.g., pneumonia, dystelectasis*; 3,1k*), Effusion* (2,5k), *Pneumothorax* (0,4k); *Central Venous Catheters* (3,0k), *Gastric Tube* (1,3k), *Thoracic Drain* (1,1k), *Tracheal Tube* (2,1k), *Misplaced Medical Device* (0,2k)]<br>(Σ: 15k)<br>**Entity Normalization:** N<br>**Annotation Guideline:** Y (see Supplement)<br>**IAA Measurement:** Y | ●<br><br>◆<br>RAD-BERT model[12] |
| ROLLER-20<br>[47] – 2020 | 118<br>+ 1,607<br>= 1,725<br>(data taken from [33]) | 90k<br>+ 68k<br>= 158k<br>(data taken from [33]) | Discharge summaries & clinical notes (nephrology) | **Annotation Types**<br>**(Annotated items)**<br>17 Named entity types<br>[*Medical condition* (11,6k), *Measurement* (5,9k), *Body part* (5,4k), *Treatment* (5,3k), *Diagnostic Procedure* (4,2k), *State of Health* (4,1k), *Process* (3,9k), *Medication* (3,5k), *Time* (3,4k), *Location* (2,1k), *Biochemistry* (1,8k), *Bioparameter* (1,6k), *Dosing* (1,3k), *Person* (1,3k), *Medical specification* (1,2k), *Medical device* (1,2k), *Body Fluid* (0,6k)] | ●<br><br>◆<br>Information extraction model[13] |

---

[12] The RAD-BERT model (trained on 3,8m on-site radiology reports and a 30k radiology-specific dictionary) is distributed via GitHub: https://github.com/rAIdiance/bert-for-radiology

[13] http://biomedical.dfki.de (this link does not direct to the language model and seems deprecated)

| | | | | | |
|---|---|---|---|---|---|
| | | | | (Σ: 58,6k)<br>10 Relation types<br>[*hasMeasure* (4,0k), *hasState* (3,5k), *isLocated* (2,9k), *hasTime* (2,4k), *Involves* (1,9k), *Shows* (1,5k), *hasDosing* (0,9k), *isSpecified* (0,8k), *Examines* (0,7k), *Severity* (0,1k)]<br>(Σ: 18,8k)<br>**Entity Normalization:** N<br>**Annotation Guideline:** N (scheme only)<br>**IAA Measurement:** N | |
| GRUNDEL-21<br>[48] – 2021 | 40,485 | noi | Discharge summaries (ophthalmology) | **Annotation Types**<br>**(Annotated items)**<br>Extraction of *Visus* (*visual acuity*; 47,6k), *Tensio* (*intraocular pressure*; 40,4k) and *Diagnoses for macular diseases* (3,2k)<br>**Entity Normalization:** Y (SNOMED-CT)<br>**Annotation Guideline:** n/a (extracted from EPR as gold standard)<br>**IAA Measurement:** n/a | ● |
| BRONCO<br>[11] – 2021 | 200<br><br>set of 11,4k shuffled sentences | 90k | Discharge summaries (oncology: hepatocellular carcinoma or melanoma) from two national hospitals (Berlin, Tübingen) | **Annotation Types**<br>**(Annotated items)**<br>*Section Headings*<br>**Entity Normalization:** N<br>**Annotation Guideline:** Y (see Supplement)<br>**IAA Measurement:** Y<br>3 named entity types<br>[*Diagnosis* (5,2k), *Treatment* (3,9k), *Medication* (2,0k)]<br>(Σ: 11,1k)<br>**Entity Normalization:** Y (ICD-10 for *Diagnosis*, OPS for *Treatment*, ATC for *Medication*)<br>**Annotation Guideline:** Y (see Supplement)<br>**IAA Measurement:** Y<br>3 types of Attributes<br>[*Laterality*: left, right, both-sided (1,3k), *Negation* (0,6k), *Speculation* (0,6k), *Possible Future Event* (0,6k)]<br>(Σ: 3,1k)<br>**Entity Normalization:** N<br>**Annotation Guideline:** Y (see Supplement)<br>**IAA Measurement:** Y | ✓<br>(DUA)[14] |
| CARDIOANNO<br>[49] – 2021 | 204<br><br>subset from ~200,000 | 382k<br><br>subset from ~218m | Discharge summaries (cardiology) | **Annotation Types**<br>**(Annotated items)**<br>12 cardiovascular concepts<br>[*angina pectoris* (0,2k), *dyspnea* (0,2k), *nycturia* (0,1k), *edema* (0,1k), *palpitation* (0,1k), *vertigo* (0,1k), *syncope* (0,2k), *arterial hypertension* (0,2k), *hypercholesterolemia* (0,1k), *diabetes mellitus* (0,1k), *familial anamnesis* (0,1k), *nicotine consumption* (0,1k)]<br>(Σ: 1,6k)<br>**Entity Normalization:** Y (ICD-10)<br>**Annotation Guideline:** Y (see Supplement)<br>**IAA Measurement:** Y | ● |

[14] https://www2.informatik.hu-berlin.de/~leser/bronco/index.html

| | | | | | |
|---|---|---|---|---|---|
| MedCorpInn<br>MedCorpInn sub<br>Karbun<br>[50,51] – 2022 | 5,003k<br>333k<br>100k | noi<br>61,117k<br>7,800k | Radiology reports | none | ● |
| Madan-22<br>[52] – 2022 | 150<br><br><br><br><br>510<br>subset from 30k | noi<br><br><br><br><br>noi | Discharge summaries (Psychiatry: Mental Status Examination (MSE) reports) | **Annotation Types**<br>**(Annotated items)**<br>*psychiatric attributes* (3,4k), *normal* (1,7k) and *pathological assessments* (1,3k), and grounding of *pathological assessments* in the AMDP terminology (1,3k)<br>(Σ: 7,7k)<br>**Entity Normalization:** Y (ICD-10 & AMDP)<br>**Annotation Guideline:** Y (see Supplement)<br>**IAA Measurement:** N<br><br>unlabelled | ● |
| [Ex4CDS]<br>[12] – 2022 | 720 | 13,4k*<br>(*estimated from averages) | Physicians' justifications supporting their estimated likelihood of future possible negative patient outcomes after transplantation<br>(kidney disease endpoints: rejection, death-censored graft loss, and infection within the next 90 days) | **Annotation Types**<br>**(Annotated items)**<br>*Risk* score [0 ... 100]<br>(Σ: 0,4k)<br>4 temporal entity types<br>[*past*, *past-to-present*, *present*, *future*]<br>12 named entity types<br>[*Condition* (1,3k), *Diagnostic Procedure* (0,1k), *Lab Value* (0,6k), *Age of Patient/Donor* (0,1k), *Medication* (0,3k), *Process* (0,2k), *Time* (0,4k), etc.]<br>(Σ: 4,2k)<br>3 relation types<br>[*hasMeasure* (0,7k), *hasState* (0,4k), *hasTimeInfo* (0,3k)]<br>(Σ: 1,4k)<br>6 factuality attributes<br>[*Positive*, *negated* (0,3k), *speculated* (0,1k), *unlikely* (< 0,1k), *minor* (< 0,1k), and *possible future* (0,1k)]<br>(Σ: 0,6k)<br>5 progression categories<br>[*risk factor*, *symptom*, *increase*, *decrease*, *conclusion*]<br>**Entity Normalization:** N<br>**Annotation Guideline:** N (scheme only)<br>**IAA Measurement:** Y | ✓ [15] |
| Roller-22<br>[53] – 2022<br>(updated version of [47]) | 61<br>+ 1,300<br>= 1,361 | 57,2k<br>+ 54,2k<br>= 111,4k | Discharge summaries & clinical notes<br>(nephrology: kidney transplantations) | **Annotation Types**<br>**(Annotated items)**<br>17 Named entity types<br>[*Medical condition* (9,0k), *Measurement* (5,4k), *Body part* (3,4k), *Treatment* (4,4k), *Diagnostic Procedure* (3,2k), *State of Health* (4,0k), *Process* (2,7k), *Medication* (3,2k), *Time* (3,1k), *Location* (1.7k), *Biochemistry* (1,4k), *Bioparameter* (1,0k), *Dosing* (1,2k), *Person* | ●<br>◆<br>Information extraction model[16]<br>(DUA) |

[15] https://github.com/DFKI-NLP/Ex4CDS

[16] https://github.com/DFKI-NLP/mEx-Docker-Deployment

| | | | | | |
|---|---|---|---|---|---|
| | | | | (1,3k), *Medical specification* (0,9k), *Medical device* (0,4k), *Body Fluid* (0,2k)]<br>(Σ: 46,4k)<br>2 Concept Attribute types<br>[*DocTime*: past, past-present, future, *Factuality*: negative, speculated, unlikely, possible future]<br>9 Relation types<br>[*hasMeasure* (3,8k), *hasState* (2,9k), *isLocated* (2,2k), *hasTime* (2,3k), *Involves* (2,0k), *Shows* (1,2k), *hasDosing* (1,2k), *isSpecified* (0,6k), *Examines* (0,4k)]<br>(Σ: 16,6k)<br>**Entity Normalization:** N<br>**Annotation Guideline:** Y (scheme only)<br>**IAA Measurement:** Y<br>PoS (STTS tag set), dependency parse trees [54] | |
| Trienes-22<br>[55] – 2022 | 851 | 327k<br>(expert)<br>463k<br>(simplified)<br>Σ: 790k | pathology reports of sarcoma patients | Parallel corpus of expert-level and layman-directed, patient-friendly parallel versions of pathology reports | ●<br>(efforts for data sharing under way) |
| Cardio:DE<br>[56] – 2023 | 500 | 993k | clinical notes and reports (cardiology: 311 in-patient & 172 out-patient letters, and 17 letters of the cardiac emergency room) | **Annotation Types**<br>**(Annotated items)**<br>14 named entity types for section headings<br>[*salutation* (0.5k), *anamnesis* (1,5k), *diagnosis* (*admission/discharge*; 9,8k), *medication* (*admission/discharge*; 7,8k), *findings* (19,3k), *lab data* (67,6k), *risk factors/allergies* (1,3k), *final recommendation* (4,5k), *summary* (3.5k), etc.]<br>(Σ: 116,9k)<br>**Entity Normalization:** Y (CDA-compliant)<br>**Annotation Guideline:** Y (see Supplement)<br>**IAA Measurement:** Y<br>2 named entity types for medication & 7 relation types<br>[*Active Ingredient* (7,6k) or *Drug* (2,1k), *Dosage* (0,2k), *Duration*(1,5k), *Form* (0,2k), *Frequency* (6,5k), *Reason* (1,5k), *Route* (0,6k), *Strength* (6,4k)]<br>(Σ: 24,2k (26,6k), with 15,1k medication relations)<br>**Entity Normalization:** N<br>**Annotation Guideline:** Y (see Supplement)<br>**IAA Measurement:** Y | ✉<br>✓<br>(DUA based on patient consent)[17] |
| Llorca-23<br>[57] – 2023 | 150<br><br><br>< 500<br><br><br>30 | 71k<br><br><br>800k<br><br><br>1,877k | Discharge summaries (oncology) from BRONCO<br>Discharge summaries (cardiology) from Cardio:DE<br>Clinical guidelines (oncology) from GGPONC 2.0 | Harmonizing approach for four German medical corpora (BRONCO, Cardio:DE, GGPONC 2.0, GraSSCo 1.0) using the BigBIO framework [58]:<br>• harmonizing different technical data formats (JSON, BRAT/BioC, etc.),<br>• harmonizing references to various terminologies (e.g., terms grounded in | ✉<br>✓<br>(DUA)[18] |

[17] https://heidata.uni-heidelberg.de/

[18] https://huggingface.co/datasets/bigbio/

| | | | | | |
|---|---|---|---|---|---|
| | (10.2k text segments)<br>63 | 43k | Synthetic discharge summaries and case reports from GRASCCO 1.0 | SNOMED CT or different versions of ICD),<br>• defining annotation mappings among "similar" named entities for entity alignment, and<br>• coping with different types of entity spans<br>**Entity Normalization:** Y (SNOMED CT, ICD-10)<br>**Annotation Guideline:** n/a<br>**IAA Measurement:** n/a | ✓<br>(public) |
| GEMTEX<br>[59] – 2023 | > 150k | | Clinical reports covering 4 medical areas<br>(cardiology, pathology, pharmacy, and neurology)<br>from 6 different clinical sites<br>(e.g., discharge summaries, findings reports) | **Annotation Types**<br>Multiple annotation layers<br>**Entity Normalization:** Y (SNOMED CT, ICD-10; planned)<br>**Annotation Guideline:** Y (not reported)<br>**IAA Measurement:** Y (not reported) | ✉<br>✓<br>(DUA based on a broad consent model) |
| 3000PA 5.0<br>[60] – 2024 | J: 1,106<br>A: 1,715<br>L: 3,823<br>Σ=6,644 | J: 1,8m<br>A: 1,7m<br>L: 3,8m<br>Σ= 7,3m | Clinical reports<br>from 3 different clinical sites (Jena, Aachen, and Leipzig) –<br>(mainly discharge summaries and transfer reports) | Automatic tagging with token and sentence boundaries (silver standard)<br>**Annotation Types**<br>**(Annotated items)**<br>Textual macrostructure segment information – section headings such as *Family & Patient Anamnesis*, *Medication*, *Diagnosis*, etc.<br>(Σ: 268k)<br>**Entity Normalization:** N (CDA-compliant)<br>**Annotation Guideline:** Y[19]<br>**IAA Measurement:** Y (not reported)<br>Named entities such as *Medications*, *Signs and Symptoms*, *Findings*, *Diagnoses*, and *PII*<br>(Σ: 1,443k)<br>**Entity Normalization:** N<br>**Annotation Guideline:** Y[20]<br>**IAA Measurement:** Y (not reported)<br>Semantic relations between named entities<br>(Σ: 135k)<br>**Entity Normalization:** N<br>**Annotation Guideline:** N<br>**IAA Measurement:** Y (not reported)<br>Temporal relations between named entities<br>(Σ: 107k)<br>**Entity Normalization:** N<br>**Annotation Guideline:** N<br>**IAA Measurement:** Y (not reported)<br>Certainty information, including negation<br>(Σ: 141k) | ● |

[19] https://doi.org/10.5281/zenodo.7707756

[20] Medications: https://doi.org/10.5281/zenodo.7707947; Signs and Symptoms, Findings, and Diagnoses: https://doi.org/10.5281/zenodo.7707917; PII: https://doi.org/10.5281/zenodo.7707882

| | | | | **Entity Normalization:** N<br>**Annotation Guideline:** N<br>**IAA Measurement:** Y (not reported)<br><br>$\Sigma_{all}$: 2,093k annotated items | |
|---|---|---|---|---|---|
| BRESSEM-24<br>[61] – 2024 | 2,000<br>+ 2,000<br>+ 2,000<br>= 6,000<br><br>subset from 3,7m radiology reports | 854k*<br>(*estimated)<br><br>520,718k | Radiology reports<br>(chest radiographs, chest CT scans, CT/radiograph examinations of the wrist covering a wide range of bone, lung, heart, and vascular diseases | **Annotation Types**<br>**(Annotated items)**<br>• presence/absence of 4 *pathologies* and 4 types of *therapy devices*,<br>• presence/absence of 23 *chest pathologies*,<br>• presence/absence of 42 named entity labels<br>**Entity Normalization:** N<br>**Annotation Guideline:** N<br>**IAA Measurement:** N | ● \| (✓) pretrained model weights for MEDBERT & radiology benchmarks)[21] |
| | | | *Additional corpus resources provided for training the medBERT model:* | | |
| | 4,369 | + 1,194k | GGPOnc 2.0 | 3 named entity types<br>[SNOMED-CT top-level hierarchies: *Finding*, *Substance*, *Procedure*]<br>($\Sigma$: 246,5k, short-span, $\Sigma$: 201,8k, long-span) | ✓(DUA) |
| | 62 | + 44k | GraSCCo | Named entity types (self-supplied)<br>($\Sigma$: 5,8k)<br>**Entity Normalization:** N<br>**Annotation Guideline:** N<br>**IAA Measurement:** N | ✓(public) |
| | 63,884 | + 12.299k | DocCheck Flexikon:<br>Open wiki about diseases, diagnostic procedures, or treatments in all areas of medicine | | ✓(public) |
| | 11,322 | + 9,324k | Webcrawl: documents from several German medical forums | | ✓(public) |
| | 12,139 | + 1,984k | German PubMed abstracts | | ✓(public) |
| | 257,999 | + 259,285k | Springer Nature: OA articles | | ✓(licenses permitting) |
| | 330,994 | + 186,201k | Thieme Publishing Group:<br>medical textbooks and journals for continuing medical education | | ✓(licenses permitting) |
| | 373,421 | + 69,639k | Electronic health records<br>from the Department of Nephrology and the Center for Kidney Transplantation at Charité: Discharge summaries & surgery reports | Codes extracted from the hospital information system<br>**Normalization:** Y (ICD-10 for diagnoses, OPS for procedures)<br>**Annotation Guideline:** n/a<br>**IAA Measurement:** n/a | ● |
| | 7,486 | + 90,381k | PhD theses from the Charité | | ✓(public) |
| | 3,639 | + 2,800k | Wikipedia: Medical entries | | ✓(public) |
| | Σ 4,723,010 | Σ 1,155,946k | | | |

[21] https://github.com/DATEXIS/medBERT.de

| | | | | | |
|---|---|---|---|---|---|
| Böhringer-24<br>[14] – 2024 | 100<br>+ 100<br>+ 100<br>= 300 | noi | ophthalmologic physicians' letters from three different German hospitals | **Annotation Types**<br>771<br>+ 1226<br>+ 809<br>= 2,806 *diagnoses*<br>(manually curated silver standard composed of ICD-10 codes)<br>**Entity Normalization:** Y (ICD-10)<br>**Annotation Guideline:** N<br>**IAA Measurement:** N | ●<br>(✉)<br>(✓)<br>(upon request)[22] |
| Idrissi-Yaghir-24<br>[62] – 2024 | 25,023k | 3,060,845k | different types of clinical reports, clinical notes, and doctor's letters | none | ● |
| RadQA<br>[62] – 2024 | 29,273<br>(question-answer pairs) | noi | question-answer pairs created from 1,223 radiology reports of brain CT scans | one custom question for every third report (covering ~400 reports)<br>**Entity Normalization:** N<br>**Annotation Guideline:** N<br>**IAA Measurement:** N | ● |
| DMP "HerzMobil"<br>[63] – 2024 | 35,579 | 1.245k*<br>(estimated from mean length) | Clinical notes | **Annotation Types**<br>**(Annotated items)**<br>9 PII types: First and last Names of Health-care professional (21,9k), Patient (16,1k), other Person (7,0k), Medical site (3,2k), Website URL (< 0,0k), Email address (~0,0k), Physical address (0,1k), Phone number (0,5k), ZIP code (0,1k)<br>($\Sigma_{all}$: 49k (silver standard))<br>**Entity Normalization:** N<br>**Annotation Guideline:** N<br>**IAA Measurement:** N | ● |
| Plagwitz-24<br>[64] – 2024 | 498 | noi | Cardiac magnetic resonance imaging (MRI) reports | **Annotation Types**<br>**(Annotated items)**<br>Attribute-value pairs for 14 *cardiac function indicators*, such as *ejection fraction* or *volumes for the left and right ventricle*<br>**Entity Normalization:** N<br>**Annotation Guideline:** N<br>**IAA Measurement:** N | ● |

**Table 1:** Real Clinical Corpora for the German Language

[22] Send requests to: daniel.boehringer@uniklinik-freiburg.de

| Corpus / Citation – Year | Docu-ments | Tokens (in 1,000=1k) | Clinical Document Types (Text Genres) | Metadata | Avail-ability<br>● Corpus<br>◆ Model |
|---|---|---|---|---|---|
| BECKER-16<br>[65] – 2016 | 61<br>+ 54<br>+ 42<br>+ 42<br>= 199 | noi | (Mixed) clinical reports:<br>discharge summaries,<br>ECG reports,<br>echo reports, and<br>radiology reports<br>(taken from the ShARe/CLEF eHealth 2013 Shared Task 1 (MIMIC II) [66] → automatic translation from English to German using Google Translate) | **Annotation Types**<br>**(Annotated items)**<br>*Disorders*<br>(Σ: 2,8k UMLS CUIs)<br>**Entity Normalization:** Y (UMLS CUIs → SNOMED-CT)<br>**Annotation Guideline:** N (re-use of ShARe/ CLEF eHealth 2013 Shared Task gold data)<br>**IAA Measurement:** N (re-use …) | ✓<br>(public) |
| N2C2-GERMAN 1.0<br>[67,68] – 2022, 2023 | 303<br>(train)<br>+ 202<br>(test)<br>= 505 | 173k<br>(train) | discharge summaries<br>[taken from the n2c2 2018 Shared Task 2 (MIMIC III) [69] → automatic translation from English to German using a pretrained neural machine translation model from *fairseq* & alignments from *fast-align* | **Annotation Types**<br>**(Annotated items)**<br>1 *Medication* entity + 6 *Medication* relation types<br>[*Drug* (8,3k) – *Strength* (4,1k), *Route* (4,5k), *Frequency* (5,2k), *Duration* (3,4k), *Form* (4,2k), *Dosage* (0,4k)]<br>(Σ: 30,2k)<br>**Entity Normalization:** N<br>**Annotation Guideline: N** (re-use of n2c2 2018 Shared Task Track gold data)<br>**IAA Measurement:** N (re-use …) | ✓<br>(public)<br>◆<br>NER model[23] |
| N2C2-GERMAN 2.0<br>[70] – 2023 | 404 | 367k | discharge summaries<br>[taken from the n2c2 2018 Shared Task 2 (MIMIC III) [69] → automatic translation from English to German using a pretrained neural machine translation model from *fairseq* & alignments from *Awesome-Align* | **Annotation Types**<br>**(Annotated items)**<br>1 *Medication* entity + 5 *Medication* relation types<br>[*Drug* (26,0k) – *Strength* (10,5k), *Frequency* (9,8k), *Duration* (1,0k), *Form* (10,5k), *Dosage* (6,7k)]<br>[Remark: *Route* (8,6k), *Reason* (6,2k), *ADE* (1,6k) were removed from the final corpus]<br>(Σ: 63.4k, without overlaps (longest span preserved); 64,5k, including overlaps)<br>(Σ: 80,9k: pre-final, full corpus)<br>**Entity Normalization:** N<br>**Annotation Guideline: N** (re-use of n2c2 2018 Shared Task Track gold data)<br>**IAA Measurement:** N (re-use …) | ✓<br>(public)<br>◆<br>NER model[24] |
| IDRISSI-YAGHIR-24<br>[62] – 2024 | noi<br>6,000k<br>(abstracts) | 695,000k<br>1,700,000 k | MIMIC III clinical notes &<br>PubMed articles<br>automatic translation from English to German using a pretrained neural machine translation model from *fairseq* | none<br>none | ◆<br>Translation -based model[25] |

**Table 2:** Translated Real Clinical Corpora for the German Language

[23] https://github.com/frankkramer-lab/GERNERMED
[24] https://github.com/frankkramer-lab/GERNERMEDpp
[25] https://huggingface.co/ikim-uk-essen

| Corpus / Citation – Year | Docu-ments | Tokens (in 1,000 = 1k) | Clinical Document Types (Text Genres) | Metadata | Avail-ability<br>● Corpus<br>◆ Model |
|---|---|---|---|---|---|
| JSynCC 1.0<br>[73] – 2018 | 399<br>+ 468<br>= 867 | 193k<br>+ 119k<br>= 313k | Operative reports<br>(orthopedics, trauma & general surgery)<br>Case reports/descriptions<br>(emergency and internal medicine, general surgery, anesthetics, ophthalmology)<br>[taken from e-book versions of medical textbooks, *manually* generated] | **Annotation Types**<br>Sentence & token splits, parts of speech (PoS)<br>**Entity Normalization:** Y (medically adapted STTS for PoS annotation)<br>**Annotation Guideline:** n/a<br>**IAA Measurement:** n/a | ✓<br>(public code base for re-building JSynCC for e-book license holders)[26] |
| GraSCCo 1.0<br>[74] – 2022 | 63 | 44k | Discharge summaries<br>[*manually* generated from real mixed-domain clinical (hospitals in Germany and Austria) and published Web resources]<br>Case reports<br>[from Open Access journals] | none | ✓<br>(public)[27] |
| Frei-23<br>[75] – 2023 | (9,845 sen-tences) | 121k | (sentences *automatically* generated via few-shot prompts (12 manually created sentences) from a large language model: *GPT NeoX* from *EleutherAI*) | **Annotation Types**<br>**(Annotated items)**<br>3 named entity types (automatically generated silver standard)<br>[*Medication* (9,9k), *Dose* (7,5k), *Diagnosis* (6,0k)]<br>(Σ: 23.4k silver items)<br>**Entity Normalization:** N<br>**Annotation Guideline:** n/a<br>**IAA Measurement:** n/a | ✓<br>(public)<br>◆<br>NER model[28] |
| JSynCC 2.0<br>[60] – 2024 | 399 | 200k | Operative reports<br>(orthopedics, trauma & general surgery)<br>Case reports/descriptions<br>(emergency and internal medicine, general surgery, anesthetics, ophthalmology)<br>[taken from e-book versions of medical textbooks, *manually* generated] | **Annotation Types**<br>**(Annotated items)**<br>Named Entities<br>[*Findings*, *Diagnoses*, *Procedures*, *PII*]<br>(Σ: 343,2k)<br>**Entity Normalization:** N<br>**Annotation Guideline:** N<br>**IAA Measurement:** N | ✓<br>(public code base for rebuil-ding JSync) |
| GraSCCo 2.0<br>[60] – 2024 | 63 | 44k | Discharge summaries<br>[*manually* generated from real mixed-domain clinical (hospitals in Germany and Austria) and published Web resources]<br>Case reports<br>[from Open Access journals] | **Annotation Types**<br>**(Annotated items)**<br>Named Entities and Semantic Relations, Temporal Relations, Certainty, Negation<br>(Σ: 177,8k)<br>**Entity Normalization:** N<br>**Annotation Guideline:** N<br>**IAA Measurement:** N | ✓<br>(public) |
| GraSCCo 3.0PHI<br>[17] – 2024 | 63 | 44k | Discharge summaries<br>[*manually* generated from real mixed-domain clinical (hospitals in Germany and Austria) and published Web resources]<br>Case reports | **Annotation Types**<br>**(Annotated items)**<br>19 PII types<br>[*Name – Patient* (0,2k), *Doctor* (0,2k), *Title* (0,1k), etc., | ✓<br>(public)[29] |

[26] Software infrastructure is available at https://github.com/JULIELab/jsyncc
[27] https://doi.org/10.5281/zenodo.6539131
[28] https://github.com/frankkramer-lab/GPTNERMED
[29] https://doi.org/10.5281/zenodo.11502329

| | | | | | |
|---|---|---|---|---|---|
| | | | [from Open Access journals] | *Date* (0,7k), *ID* (0,1k)<br>Location – *City* (< 0,1k), *ZipCode* (0,1k), *Street* (< 0,1k), *Hospital* (< 0,1k),<br>Contact – *Phone* (< 0,1k), etc.]<br>(Σ: 1,4k)<br>**Entity Normalization:** N<br>**Annotation Guideline:** Y[19]<br>**IAA Measurement:** Y | |

**Table 3:** Synthetic Clinical Corpora for the German Language

| CORPUS / Citation – Year | Docu-ments | Tokens (in 1000 = 1k) | Medical Document Types (Text Genres) | Metadata | Avail-ability<br>● Corpus<br>◆ Model |
|---|---|---|---|---|---|
| BROWN-02<br>[80] – 2002 | 531,690 (journal article titles) | ~ [4,000-5,000]k | Parallel corpus (English–German) of paired journal article titles retrieved from PubMed | none | ✓ |
| MUCHMORE<br>[81] – 2002 | ~ 9,000 (abstracts for each language) | ~ 1,000k | Parallel corpus (English–German) of abstracts from 41 medical journals hosted at the Springer Web site covering various medical sub-domains (e.g. neurology, radiology) | **Annotation Types**<br>Sentences, tokens, parts of speech (PoS), morphological segmentation, phrasal chunks<br>(automatically generated silver standard)<br>**Lexical Normalization:** (UMLS Specialist Lexicon)<br>**Annotation Guideline:** n/a<br>**IAA Measurement:** n/a<br>term mapping to MeSH codes (subset of UMLS Metathesaurus) & semantic relations from the UMLS Semantic Network<br>(automatically generated silver standard)<br>**Entity Normalization:** Y (UMLS & EuroWordNet)<br>**Annotation Guideline:** n/a<br>**IAA Measurement:** n/a | ✓ |
| SPRINGER-LINK<br>[82] – 2003 | 5,271 | ~ 910k | Abstracts of German medical journal publications, available from an online library for medicine (SpringerLink) | **Annotation Types**<br>Automatically derived index terms<br>**Entity Normalization:** Y (local dictionary linked with MeSH terms)<br>**Annotation Guideline:** n/a<br>**IAA Measurement:** n/a | ✓ |
| SPRINGER<br><br>MEDTITLE<br>[83] – 2004 | 9,640 | ~ 450k [30k sentences]<br>~ 5,500k [549k sentences] | titles plus abstracts of medical journal articles from Springer, each in German (& in English);<br>paired titles of medical journal articles (from PubMed) | none | ✓ |
| FRAMED<br>[21] – 2004 | noi (~6,500 sentences) | 100k | Various clinical report types (discharge, pathology, histology, and surgery reports),<br>a medical textbook, and<br>Web documents taken from a consumer health care portal (netdoktor) | **Annotation Types**<br>Sentence & token splits, parts of speech (PoS)<br>**Entity Normalization:** Y (medically adapted STTS for PoS annotation)<br>**Annotation Guideline:** N<br>**IAA Measurement:** Y | ●<br>◆<br>FRAMED model as part of JCORE [14,15] |
| MORIN-12<br>[84] – 2012 | 103<br>118 (English)<br>130 (French) | 220k<br>265k (English)<br>265k (French) | Multilingual comparable corpus (English, French, German) from scientific paper websites, with hits for *"breast cancer" ('cancer du sein'* in French and *'Brustkrebs'* in German) in titles & keyword sections only | none | ✓ |
| MANTRA [SILVER]<br>[85] – 2014 | $\Sigma_{EFGSD}$:<br>4,255k<br><br>719k<br>+ 141k<br>+ 121k<br>= 981k | $\Sigma_{EFGSD}$:<br>60,424k<br><br>5,997k<br>+ 2,100k<br>+ 5,194k<br>=13,291k | Multilingual parallel corpus: English, French, German, Spanish, Dutch), including<br>Medline titles (PubMed)<br>Drug labels (EMEA)<br>Patent claims (EPO) | **Annotation Types**<br>**(Annotated items)**<br>Automatically generated named entities from ensembles of NER taggers (silver standard) – mapped to UMLS CUIs and semantic groups, such as *Activities & Behaviors*, *Anatomy, Chemicals & Drugs, Devices, Disorders, Geographic Areas, Living Beings, Objects, Phenomena, Physiology*<br>(Σ: 75,2k for German; harmonized by threshold)<br>($\Sigma_{all}$: > 221k, in total)<br>**Entity Normalization:** Y (UMLS – MeSH, SNOMED CT, MedDRA)<br>**Annotation Guideline:** n/a<br>**IAA Measurement:** n/a | ✓ |
| MANTRA GSC<br>[86] – 2015 | $\Sigma_{EFGSD}$:<br>1,450<br><br>+ 100<br>+ 100<br>+ 50<br>= 250<br><br>(Subset of [85]) | $\Sigma_{EFGSD}$:<br>29,329<br><br>947<br>+ 1,956<br>+ 3,117<br>= 6,020<br><br>(Subset of [85]) | Multilingual parallel corpus: English, French, German, Spanish, Dutch), including<br>Medline titles (PubMed)<br>Drug labels (EMEA)<br>Patent claims (EPO) | **Annotation Types**<br>**(Annotated items)**<br>named entities (gold standard) – mapped to UMLS CUIs and semantic groups, such as *Anatomy, Chemicals & Drugs, Devices, Disorders, Geographic Areas, Living Beings, Objects, Phenomena, Physiology, Procedures*<br>(Σ: 1,082 for German)<br>($\Sigma_{all}$: 5,530, in total)<br>**Entity Normalization:** Y (UMLS – MeSH, SNOMED CT, MedDRA)<br>**Annotation Guideline:** Y (see Supplement; broken link)<br>**IAA Measurement:** Y | ✓ |
| HIML 1.0<br>[87] – 2017 | 781k<br><br>+ 33k<br>+ 1,848k<br>= 2,662k | > 60,000k<br>(estimated) | Multilingual parallel corpus (English, German), including<br>EMEA (European Medicines Agency) documents<br>MUCHMORE segments<br>MAREC patent documents | none | ✓<br>(upon request) |
| EFSG-UVIGOMED<br><br>ML–UVIGOMED<br>[88] – 2018 | 2,130<br>$\Sigma_{all}$: 19,210<br><br>3,147<br>$\Sigma_{all}$: 23,647 | ~ 500k | Multi-lingual corpus: MEDLINE/PUBMED abstracts (English, French, Spanish, German) about 26 types of *Diseases*<br><br>WIKIPEDIA articles (German, English, French, Spanish, Italian, Galician, Romanian, Slovene, and Icelandic) about Human Medicine (including 22 subcategories, such as Cardiology, Endocrinology, Human Genetics, Geriatrics, Nephrology, | Index terms extracted from Medline<br>**Entity Normalization:** Y (MeSH)<br>**Annotation Guideline:** n/a<br>**IAA Measurement:** n/a | ✓<br><br>✓ |

| | | | Neurology, Oncology, Surgery and Urology, Rheumatology) | | |
|---|---|---|---|---|---|
| VILLENA-20<br>[89] – 2020 | 59,539<br>$\Sigma_{all}$: 93,969 | 20,438k<br>$\Sigma_{all}$: 83.869k | (Web-scraped) Multilingual corpus (German, English, Spanish), with a 63% share of German-language medical full-text articles/abstracts | none | ✓<br>Zenodo[30] |
| GGPONC 1.0<br>[90] – 2020 | 25<br>(4.2k annotated text segments)<br><br>Subset of 8.4k text segments | 664k<br><br>Subset of 1,340k | (all) Clinical Practice Guidelines of the German Cancer Society (oncology) | **Annotation Types**<br>**(Annotated items)**<br>7 named entity types<br>[UMLS Semantic Groups: *Anatomical Structure, Chemicals & Drugs, Devices, Disorders, Living Being, Physiology, Procedures*]<br>1 attribute type: *TNM*<br>[silver standard, manually validated]<br>(Σ: 73,8k) [91]<br><br>**Entity Normalization:** Y (UMLS)<br>**Annotation Guideline:** N<br>**IAA Measurement:** Y<br><br>plus evidence-based recommendation meta-data, such as *Type of Recommendation, Recommendation Grade, Strength of Consensus, Expert Opinion, Level of Evidence, Literature References*, etc. | ✓<br>(DUA)[31] |
| GGPONC 2.0<br>[91] – 2022 | 30<br>(5k annotated text segments)<br><br>(Subset of 10.2k text segments)<br><br>(Superset of [90]) | 830k *<br>(estimated)<br><br>Subset of 1,877k | (all) Clinical Practice Guidelines of the German Cancer Society (oncology) | **Annotation Types**<br>**(Annotated items)**<br>3 named entity types<br>[SNOMED-CT top-level hierarchies:<br>*Finding* (132,8k ss\| 105,0k ls; Diagnosis or Pathology, Other Finding),<br>*Substance* (24,9k ss \| 18,2k ls; Clinical Drug, Nutrient/ Body Substance, External Substance),<br>*Procedure* (88,9k ss \| 77,7k ls; Therapeutic, Diagnostic)]<br>(Σ: 246,5k, short-span (ss),<br>Σ: 201,8k, long-span (ls))<br>**Entity Normalization:** Y (SNOMED-CT)<br>**Annotation Guideline:** Y[32]<br>**IAA Measurement:** Y | ✓<br>(DUA)[33] |
| BTC<br>[53] – 2022 | noi<br>(~7,7GB) | noi | Web documents taken from a consumer health care portal &<br>Medical newspapers &<br>(German) PubMed abstracts &<br>Clinical case studies &<br>Medical textbooks | none | ✓ |
| CHADL<br>[92] – 2022 | 50 | 32k | Discharge summaries (neurology)<br>[because of the small number of tokens & documents the clinical portion of this corpus is excluded from deeper consideration] | **Annotation Types**<br>Section Headings (8 categories)<br>[*Header and Footer, Personal Data, Diagnoses, Anamneses, Medication, Procedures & Measures, Findings, Epicrisis*] | ✓<br>(access is granted to institutions adhering to trusted data |

[30] https://zenodo.org/record/3463379.XY4RsUEzaV4
[31] https://www.leitlinienprogramm-onkologie.de/projekte/ggponc-english/
[32] https://github.com/hpi-dhc/ggponc_annotation
[33] https://www.leitlinienprogrammonkologie.de/projekte/ggponc-english/

| | | | | | |
|---|---|---|---|---|---|
| | | | | 4 named entity types<br>*[Medication – Dosage, Intake (medication order), Disorder, (therapeutic) Procedures, Diagnostic Measures]*<br>**Entity Normalization:** N<br>**Annotation Guideline:** Y (see Supplement)<br>**IAA Measurement:** Y | privacy policies and protocols) |
| | | 7,069k | Drug labels | none | ✓ |
| | | + 38,374k | Bio-medical abstracts (LIVIVO) | none | ✓ |
| | | + 20,637k<br>= 66,080k | Medical WIKIPEDIA articles | none | ✓ |
| BRESSEM-24<br>[61] – 2024 | 2,000<br>+ 2,000<br>+ 2,000<br>= 6,000<br><br>subset from 3,7m radiology reports | 854k*<br>(*estimated)<br><br>520,718k | Radiology reports<br>(chest radiographs, chest CT scans, CT/radiograph examinations of the wrist covering a wide range of bone, lung, heart, and vascular diseases | **Annotation Types**<br>**(Annotated items)**<br>• presence/absence of 4 *pathologies* and 4 types of *therapy devices* &<br>• presence/absence of 23 *chest pathologies* &<br>• presence/absence of 42 named entity labels<br>**Entity Normalization:** N<br>**Annotation Guideline:** N<br>**IAA Measurement:** N | ● \| (✓) pretrained model weights for MEDBERT) |
| | | | *Additional corpus resources for training the medBERT model:* | | |
| | 4,369 | + 1,194k | GGPOnc 2.0 | 3 named entity types<br>[SNOMED-CT top-level hierarchies: *Finding*, *Substance*, *Procedure*]<br>(Σ: 246,5k, short-span, Σ: 201,8k, long-span) | ✓(DUA) |
| | 62 | + 44k | GraSCCo | Named entity types (self-supplied)<br>(Σ: 5,8k)<br>**Entity Normalization:** N<br>**Annotation Guideline:** N<br>**IAA Measurement:** N | ✓(public) |
| | 63,884 | + 12.299k | DocCheck Flexikon:<br>Open wiki about diseases, diagnostic procedures, or treatments in all areas of medicine | | ✓(public) |
| | 11,322 | + 9,324k | Webcrawl: documents from several German medical forums | | ✓(public) |
| | 12,139 | + 1,984k | German PubMed abstracts | none | ✓(public) |
| | 257,999 | + 259,285k | Springer Nature: OA articles | none | ✓(licenses permitting) |
| | 330,994 | + 186,201k | Thieme Publishing Group:<br>medical textbooks and journals for continuing medical education | none | ✓(licenses permitting) |
| | 373,421 | + 69,639k | Electronic health records<br>from the Department of Nephrology and the Center for Kidney Transplantation at Charité: Discharge summaries & surgery reports | Codes extracted from the hospital information system<br>**Normalization:** Y (ICD-10 for diagnoses, OPS for procedures) | ● |
| | 7,486<br>3,639<br>Σ 4,723,010 | + 90,381k<br>+ 2,800k<br>Σ 1,155,946k | PhD theses from the Charité<br>Wikipedia: Medical entries | **Annotation Guideline:** n/a<br>**IAA Measurement:** n/a<br>none | ✓(public)<br>✓(public) |
| IDRISSI-YAGHIR [62] – 2024 | ~ 6,000k<br>6,000k (abstracts) | 695,000k<br>1,700,000 k | MIMIC III clinical notes &<br>PubMed articles<br>automatic translation from English to German using a pretrained neural machine translation model from *fairseq* | none<br>none | ◆<br>Translation-based model[34] |

**Table 4:** Close Domain Proxies: Pseudo-Clinical Corpora for the German Language

[34] https://huggingface.co/ikim-uk-essen

| Corpus / Citation – Year | Documents | Tokens (in 1000) | Medical Document Types (Text Genres) | Metadata | Availability<br>● Corpus<br>◆ Model |
|---|---|---|---|---|---|
| FraMed<br>[21] – 2004 | noi<br>(~6,500 sentences) | 100k | Various clinical report types (discharge, pathology, histology, and surgery reports),<br>a medical textbook, and<br>Web documents taken from a consumer health care portal (netdoktor) | **Annotation Types**<br>Sentence & token splits, parts of speech (PoS)<br>**Entity Normalization:** Y (medically adapted STTS for PoS annotation)<br>**Annotation Guideline:** N<br>**IAA Measurement:** Y | ●<br>◆<br>FraMed model as part of JCoRe [22,23] |
| Lohr-16<br>[31] – 2016 | 450<br>subset from 22,4k | 266k | Operative reports (digestive tract) | **Annotation Types**<br>*Diagnoses, Procedures*<br>**Entity Normalization:** Y (ICD for diagnoses, OPS for executed procedures)<br>**Annotation Guideline:** n/a (extracted from EPR as gold standard)<br>**IAA Measurement:** n/a | ● |
| | 5,8m | 125,9m | (Fragments of) newspaper articles with medical content extracted from DWDS (*Digitales Wörterbuch der Deutschen Sprache*) | Mentions of 400 medical terms, such as *"patient", "surgery", "ambulance"*, etc.<br>**Entity Normalization:** N<br>**Annotation Guideline:** n/a<br>**IAA Measurement:** n/a | ✓ |
| EFSG-UVigoMED | 2,130<br>$\Sigma_{all}$: 19,210 | ~ 500k | Multi-lingual corpus:<br>Medline/PubMed abstracts (German, English, French, Spanish) about 26 types of *Diseases* | Index terms extracted from Medline<br>**Entity Normalization:** Y (MeSH)<br>**Annotation Guideline:** n/a<br>**IAA Measurement:** n/a | ✓ |
| ML–UVigoMED<br>[88] – 2018 | 3,147<br>$\Sigma_{all}$: 23,647 | | Wikipedia articles (German, English, French, Spanish, Italian, Galician, Romanian, Slovene, and Icelandic) about Human Medicine (incl. 22 subcategories, such as Cardiology, Endocrinology, Human Genetics, Geriatrics, Neurology, Nephrology, Oncology, Rheumatology, Surgery, Urology) | Wikipedia categories related to Human Medicine | ✓ |
| WikiSection<br>[93] – 2019 | 2,3k<br>(Diseases, German only)<br>subset from 38k | ~2,000k*<br>(*estimate)<br>(45.7 sentences/ article) | Wikipedia articles (German, English) about *Diseases* (and *Cities*) | **Annotation Types**<br>**(Annotated items)**<br>25 topic classes<br>[*Diagnosis, Treatment, Symptoms, Mechanism, Medication, Classification,* etc.] for 6,1k headings<br>**Entity Normalization:** Y (Wikidata Categories (for topic classes) & BabelNet synsets (for headings))<br>**Annotation Guideline:** N<br>**IAA Measurement:** Y<br>$\Sigma_{all}$: 242k labeled sections and normalized topic labels for up to 30 topics | ✓[35] |

[35] https://github.com/sebastianarnold/WikiSection

| | | | | | |
|---|---|---|---|---|---|
| TLC-MED1<br>[94] – 2020 | 2k<br>(kidney diseases)<br>2k<br>(stomach and intestines)<br>(Σ: 4k) | 204k<br>(kidney diseases)<br>235k<br>(stomach and intestines)<br>(Σ: 439k) | Threads from the German-language patient forum MED1 | **Annotation Types**<br>**(Annotated items)**<br>Paraphrase equivalence links between medical expert (Σ: 1,7k) and medical layman expressions (Σ: 4,7k), with focus on *Symptoms, Diseases, Treatments & Examinations*<br>**Entity Normalization:** (UMLS & Wiktionary)<br>**Annotation Guideline:** N<br>**IAA Measurement:** N | ✓ |
| RSS<br>[95] – 2020 | noi | 13,649k | RSS feeds about the corona-virus pandemic from 13 German newspapers and 3 non-print outlets:<br>print: Focus Online, Frankfurter Allgemeine Zeitung, Frankfurter Rundschau, Süddeutsche Zeitung, Neue Zürcher Zeitung, SpiegelOnline, Standard, tageszeitung (TAZ), Die Welt, and Die Zeit;<br>non-print: web.de, t-online.de,& heise.de | COWIDPLUS ANALYSIS generates lexicographic metadata from RSS:<br>• daily and weekly frequency lists of token unigrams (POS-tagged and lemmatized) and bigrams,<br>• daily values for the central corpus measures (redundancy, mean segmental type-token ratio (MSTTR), & top 100 accumulated token frequency share | (✓)<br>(metadata only) |
| BECK-21<br>[96] – 2021 | 3k<br>subset from 238k | (~555k*)<br>(*estimate) | Tweets<br>(selected by search terms, such as *Corona, Pandemic, Covid 19, Social distance*, etc.) | **Annotation Types**<br>**(Annotated items)**<br>4-category label system indicating the tweet's stance towards governmental measures taken against the pandemic<br>[*Refute (negative; 0,3k), Support (positive; 0,5k), Comment (neutral; 1,1k), Unrelated (no measures mentioned; 1,0k)*]<br>(Σ: 3.0k)<br>**Entity Normalization:** N<br>**Annotation Guideline:** Y (see Appendix)<br>**IAA Measurement:** Y | ✓ |
| FANG-COVID<br>[97] – 2021 | 28,1k<br>+ 13,2k<br>= 41,3k<br>(complete news article / tweet) | 22,000k*<br>+ 10,600k*<br>= 32,600k*<br>(*estimate) | Real news articles and tweets<br>Fake news articles and tweets<br>(selected by the query terms: *Corona, Covid, Infektion, Lockdown, Impfen, Impfung, Impfstoff*) | Automatically generated meta information relating to the articles' spreading on social media (e.g., likes, quotes, retweets, replies) and user characteristics (e.g., number of followers & friends) | ✓[36] |
| LIFELINE 1.0<br>[98] – 2022 | 101<br>(complete forum post)<br>subset from 4,169 | 11,k<br>463k*<br>(*estimate) | Threads about Adverse Drug Reactions (ADRs) from the German-language patient forum LIFELINE | **Annotation Types**<br>**(Annotated items)**<br>(Binary) categorization of documents into those reporting ADRs (101 posts) and non-ADR ones (4068 posts)<br>(Σ: 4.2k)<br>**Entity Normalization:** N<br>**Annotation Guideline:** Y<br>**IAA Measurement:** N | ✓<br>(DUA)[37] |

[36] https://github.com/justusmattern/fang-covid

[37] https://github.com/DFKI-NLP/cross-ling-adr

| | | | | | |
|---|---|---|---|---|---|
| BTC<br>[53] – 2022 | noi<br>(~7,7GB) | noi | Web documents taken from a consumer health care portal & Medical newspapers & (German) PubMed abstracts & Clinical case studies & Medical textbooks | none | ✓ |
| CHADL<br>[92] – 2022 | 50 | 32k | Discharge summaries (neurology)<br>[because of the small number of tokens & documents the clinical portion of this cor-pus is excluded from deeper consideration] | **Annotation Types**<br>Section Headings (8 categories)<br>[*Header and Footer, Personal Data, Diagnoses, Anamneses, Medication, Procedures & Mea-sures, Findings, Epicrisis*]<br>4 named entity types<br>*[Medication – Dosage, Intake (medication order), Disorder, (therapeutic) Procedures, Diagnostic Measures]*<br>**Entity Normalization:** N<br>**Annotation Guideline:** Y (see Supplement)<br>**IAA Measurement:** Y | ✓<br>(access is granted to institutions adhering to trusted data privacy policies and protocols) |
| | | 7,069k | Drug labels | none | ✓ |
| | | + 38,374k | Bio-medical abstracts (LIVIVO) | none | ✓ |
| | | + 20,637k<br>= 66,080k | Medical WIKIPEDIA articles | none | ✓ |
| BRESSEM-24<br>[61] – 2024 | 2,000<br>+ 2,000<br>+ 2,000<br>= 6,000 | 854k*<br>(*estimated) | Radiology reports<br>(chest radiographs, chest CT scans, CT/radiograph examinations of the wrist covering a wide range of bone, lung, heart, and vascular diseases) | **Annotation Types**<br>**(Annotated items)**<br>• presence/absence of 4 *pathologies* and 4 types of *therapy devices* &<br>• presence/absence of 23 *chest pathologies* &<br>• presence/absence of 42 named entity labels<br>**Entity Normalization:** N<br>**Annotation Guideline:** N<br>**IAA Measurement:** N | ● \|<br>(✓)<br>pretrained model weights for MEDBERT) |
| | subset from 3,7m radiology reports | 520,718k | | | |
| | | | *Additional corpus resources for training the medBERT model:* | | |
| | 4,369 | + 1,194k | GGPOnc 2.0 | 3 named entity types<br>[SNOMED-CT top-level hierarchies: *Finding*, *Substance*, *Procedure*]<br>(Σ: 246,5k, short-span, Σ: 201,8k, long-span) | ✓(DUA) |
| | 62 | + 44k | GraSCCo | Named entity types (self-supplied)<br>(Σ: 5,8k)<br>**Entity Normalization:** N<br>**Annotation Guideline:** N<br>**IAA Measurement:** N | ✓(public) |
| | 63,884 | + 12.299k | DocCheck Flexikon:<br>Open wiki about diseases, diagnostic procedures, or treatments in all areas of medicine | none | ✓(public) |
| | 11,322 | + 9,324k | Webcrawl: documents from several German medical forums | none | ✓(public) |
| | 12,139 | + 1,984k | German PubMed abstracts | | ✓ (public) |
| | 257,999 | + 259,285k | Springer Nature: OA articles | | ✓ (licenses permitting) |
| | 330,994 | + 186,201k | Thieme Publishing Group:<br>medical textbooks and journals for continuing medical education | | ✓ (licenses permitting) |
| | 373,421 | + 69,639k | Electronic health records<br>from the Department of Nephrology and the Center for Kidney Transplantation at Charité:<br>Discharge summaries & surgery reports | Codes extracted from the hospital information system<br>**Normalization:** Y (ICD-10 for diagnoses, OPS for procedures)<br>**Annotation Guideline:** n/a<br>**IAA Measurement:** n/a | ● |
| | 7,486 | + 90,381k | PhD theses from the Charité | none | ✓ (public) |
| | 3,639 | + 2,800k | Wikipedia: medical entries | none | ✓ (public) |
| | Σ 4,723,010 | Σ 1,155,946k | | | |
| LIFELINE 2.0 [99][38] – 2024 | 118<br>(complete forum post)<br>subset from ~10k | 29,0k | Threads about Adverse Drug Reactions (ADRs) from the German-language patient forum LIFELINE (comparable data also available for French & Japanese) | **Annotation Types**<br>**(Annotated items)**<br>12 entity types, 4 attribute types, and 13 relation types related to ADRs:<br>Entities and associated attributes, e.g.,<br>• *Drug* (0,6k), with attributes *increase, de-crease, stopped, started, unique_dose,*<br>• *Time*, with attributes *frequency, duration, date, point in time,*<br>• *Disorder* (1,2k), *Route, Anatomy, (Body) Function, Test,* etc.<br>(Σ: 3,5k entities and 1,1k attributes)<br>($\Sigma_{ent+att}$: 4,6k)<br>**Entity Normalization:** N<br>**Annotation Guideline:** Y[39]<br>**IAA Measurement:** Y<br>Relations and associated entities, e.g.:<br>• *Caused*: drug OR disorder, dis-order OR (body) function<br>• *Treatment_for*: drug, disorder OR (body) function<br>• *Has_dosage*: drug, measure<br>• *Has_result*: test, measure OR disorder OR (body) function<br>• *Examined_with*: disorder OR Anatomy OR (body) function, test<br>• *Interacted_with*: drug, drug<br>• *Has_route*: drug,route<br>(Σ: 2,2k)<br>**Entity Normalization:** N<br>**Annotation Guideline:** Y[27]<br>**IAA Measurement:** Y<br>($\Sigma_{all}$:6,8k) | ✓<br>(DUA) |

[38] LIFELINE 2.0 contains a different set of documents than LIFELINE 1.0 (thus, they are counted as two distinct corpora) although the theme covered remains the same.

[39] https://github.com/DFKI-NLP/keepha_annotation_guidelines/blob/main/KEEPHA_annotation_guidelines.pdf

| | | | | | |
|---|---|---|---|---|---|
| | | | | (Binary) categorization of documents into those reporting ADRs (originally, 324 posts; 118 posts after additional (length) filtering) and non-ADR ones (9,7k posts)<br>**Entity Normalization:** N<br>**Annotation Guideline:** N<br>**IAA Measurement:** N | |
| HEINRICH-24 [100] – 2024 | 1099 (posts)<br>Subset from > 13 million posts collected from over 200 different Telegram channels | ~198k<br>Subset from ~ 400 million tokens | Posts from Telegram on conspiracy narratives surrounding the COVID-19 pandemic | **Annotation Types**<br>**(Annotated items)**<br>14 labels for conspiracy-related or conspiracy-adjacent content,<br>[e.g., pseudo-pandemic, criticism of countermeasures, alternative treatments, vaccine hazards, COVID-19 conspiracies, other conspiracies, QAnon, group-focused enmity, state as an enemy, indoctrination, esotericism & pseudo-science, etc.]<br>(Σ: ~ 0,8k)<br>**Entity Normalization:** N<br>**Annotation Guideline:** Y (not reported)<br>**IAA Measurement:** Y | ✓<br>(DUA)[40] |
| HEALTHFC [101] – 2024 | 750 (health-related claims & evidence information) | ~ 675k | Bilingual corpus (English – German) for medical fact checking selected from the Web portal *Medizin Transparent* | **Annotation Types**<br>**(Annotated items)**<br>*Claim – evidence – verdict* text triples:<br>(Public health) *claims* automatically selected from the Web portal<br>(related, e.g., to eating habits, dietary topics, the immune system, the respiratory, musculo-skeletal, or cardiovascular systems, alternative medicine, etc.),<br>*evidence* sentences for each claim<br>(manually extracted from clinical trials or systematic reviews that were manually phrased in layman language and manually annotated, incl. medical explanations),<br>*verdicts* manually assembled from medical experts<br>(i.e., *supported* (202), *refuted* (125), *not enough information* (423))<br>(Σ: 750 triples)<br>**Entity Normalization:** N<br>**Annotation Guideline:** N<br>**IAA Measurement:** Y | ✓<br>(Github)[41] |
| PEDRINI-24 [102] – 2024 | 60 (CT summaries re-phrased in layperson language) | 145k | Parallel corpus (English, German, Italian, so altogether 180 CT summaries) of layperson summaries of clinical trials (CT) | none | ✓ |

[40] https://corpora.linguistik.uni-erlangen.de/cqpweb/schwurpus_v2/ and https://github.com/fau-klue/infodemic
[41] https://github.com/jvladika/HealthFC/

| Frei-24<br>[103] – 2024 | 84,478<br>(text fragments) | 2,023k | Wikipedia text fragments, labelled with an *Anatomical Therapeutic Chemical* (ATC) code | **Annotation Types**<br>**(Annotated items)**<br>ATC code tags (automatically extracted from WikiData)<br>(Σ: 105,2k codes)<br>**Entity Normalization:** Y (WikiData QID numbers & ATC)<br>**Annotation Guideline:** n/a<br>**IAA Measurement:** n/a | ✓<br>(Git-hub)[42] |
|---|---|---|---|---|---|

**Table 5:** Distant-Domain Proxies: Medical Non-Clinical Corpora for the German Language

[42] https://github.com/frankkramer-lab/WikiOntoNERCorpus

**APPENDIX**

***Clinical/Medical Corpus Documentation: Corpus Card***

Abstracting away from the particularities of individual corpora in the papers we screened for this review, we here propose a general template datasheet for corpus descriptions, we call *Corpus Card*. Table 7 summarizes a generalized set of mandatory and (desirable) optional attributes we deem important for proper and informative corpus documentation.

| Corpus Attributes | Definition | ***Attribute Values (in italics)*** **& Examples** |
|---|---|---|
| **Language(s)** | Natural language(s) of the document units in the corpus | • *de*: German<br>• *en*: English<br>• *es*: Spanish<br>• *fr*: French, etc.<br>[Codes based on ISO 639] |
| **Modality** | The mode how natural language utterances are communicated or overlaid | • *written* language (textual documents)<br>• *spoken* language (recorded speech, non-transcribed audio signals)<br>• *visual* signals, mostly body movements (gestures, face signals, deictic moves, etc.)<br>• *multi-modal* – a mixture of different modalities |
| **Media** | Types of media (data) complementing natural language utterances | • *visual* data (images, photos, drawings, diagrams, charts, figures, movies, etc.)<br>• *structured* data (tables, etc.)<br>• *non-speech* auditory data (music, sounds, etc.)<br>• *sensor* signals (measured physiological data, click streams, social networking data, etc.)<br>• *multi-media* – a mixture of various media |
| **Data status** | Status of the data, i.e., whether they are originally taken from the domain of discourse, or whether they are purposefully altered/modified (hence, not original) | • *original* (i.e., authentic, de-identified) data<br>• *translated* data (mostly automatic language-to-language translations, e.g., en2de)<br>• *synthetic* (i.e., fictitious, invented) data |
| **Corpus Versioning** | If a family of versions of basically the same or a similar corpus emerges, the relationship of a specific corpus to its predecessors should be made explicit in set-theoretical terms | • = (the current corpus version shares all documents with a specified reference version)<br>• *!=* (the current corpus version shares no document with a specified reference version)<br>• *SuperSet/SubSet-of* (the current corpus version is a superset/subset of the documents of a specified reference version)<br>• *ProperIntersect* (the current corpus version has a non-empty intersection with the documents of a specified reference version, yet is neither a subset nor a superset) |

| **# Documents** | Absolute number of document units | |
|---|---|---|
| **# Superset** | The (size of the) superset from which a subset (= corpus) was drawn, if any | Comment: # Superset >> # Documents |
| **# Tokens** | Absolute number of single “words” | |
| # Types | Absolute number of distinct single “words” | |
| # Other document units | Absolute number of sentences, paragraphs, sections/chapters, segments (in parallel/ comparable corpora), etc., if any | |
| Average length of documents | Arithmetic *mean* (incl. standard deviation) or *median* of #tokens (or other document units, if any) per document | |
| Sampling strategy | How were the documents sampled? | • *ad hoc* sample (arbitrary, often subjective selection of data items)<br>• *random* sample, etc. |
| Data splits | (Recommended) data splits for training, development/validation, testing in the corpus | An example:<br>70:15:15 |
| **Release conditions** | Distribution status of the corpus, i.e., whether the corpus is publicly sharable or not and, if so, under which access conditions | • *Non-distributable*, classified, inaccessible for external use<br>• *Regulated* distribution on a contractual basis (e.g., DUA, IPR licence, royalties/fees)<br>• *Publicly* shared (e.g., on online sites such as Zenodo, GitHub, etc.), without constraints |
| Technical format | Storage format of the corpus | • UTF-8, XML, JSON, etc. |
| **Contact Data** | For *regulated* access: The digital address of the person in charge of the corpus (the corpus owner or administrator)<br>For *public* access: the digital address of the site which hosts the corpus | • An *e-mail* address<br>• A *URI* (URL or URN) |
| **Genre Attributes** | | |
| **Verbal interaction mode** | Types of verbal interaction modes prevalent in the document units of the corpus | • *monologic* data: typically, written texts, with readers as addressees, such as clinical reports or notes, encyclopedia articles, scientific papers/abstracts or newspaper articles, books, etc.<br>• *dialogic* data: typically, written or spoken utterance exchanges between two speakers, such as tweets, chats, posts, question-answer sequences, conversations<br>• *multi-party* data: typically, written or spoken utterance exchanges between more than two speakers, e.g., in meetings, discussion groups |
| **(Medical) Document genre(s)** | Types of (medical) documents characterized by normative writing habits, conventions relating to contents, communicative goals and formal document structure, as well as type-specific linguistic style relating to choic- | Monologic:<br>• *clinical reports*/notes, such as discharge summaries, pathology or radiology reports, nurse notes, case reports, etc.<br>• *clinical guidelines* |
| | es of terminology, abbreviations & acronyms, phrasal patterns, etc. | • *clinical trial* reports<br>Dialogic:<br>• *patient-doctor* conversations<br>Multi-party:<br>• oncologic *council*<br>Comment: For reasons of clarity and added value,<br>• *clinical domains*, such as vascular and casualty surgery, internal medicine, neurology, anaesthesia, intensive care, radiology, physiotherapy, and<br>• *anatomical regions* targeted in the documents, as with lung cancer, thorax X-rays, etc.<br>should be co-mentioned with medical document genres |
| # Documents/genre | Absolute number of documents per genre | |
| Average length of documents/genre | Arithmetic *mean* (incl. standard deviation) or *median* of #tokens per genre | |
| **Institutional Attributes** | | |
| # Clinical sites/ institutions | Absolute number of clinical sites from each clinical institution (represented in the corpus) | An example:<br>Intensive Care Unit, Children's Hospital, Neurosurgery Dept. @ Mayo Clinic Hospital<br>→ 3 clinical sites, 1 institution |
| # Clinical institutions | Absolute number of clinical institutions (represented in the corpus) | An example:<br>Mayo Clinic Hospital, The Vanderbilt Clinic – Nashville, Kerrville VA Hospital<br>→ 3 institutions |
| # Countries / Languages | Absolute number of countries and languages (represented in the corpus) | An example:<br>Mayo Clinic Hospital, USA; Klinikum rechts der Isar, Germany<br>→ 2 countries / 2 languages |
| **Metadata Attributes** | | |
| **Annotation types (& attributes)** | Clinically relevant metadata categories: Boolean and multi-valued categories, named entities and associated attributes, relations, etc. | An example:<br>Symptom, finding, diagnosis<br>→ 3 named entity types<br>Drug: frequency, dosage, mode, duration<br>→ 1 named entity type, 4 attributes<br>Smoker status: Smoker/non-smoker<br>→ 1 categorical type (Boolean)<br>Age: infant, adolescent, adult, elderly<br>→ 1 categorical type (4-valued)<br>Treatment_for: drug, disorder<br>Has_result: test, (body) function<br>Interacted_with: drug, drug<br>Time_before: disorder, disorder<br>→ 4 relation types |
| **# Annotation in-stances/annotation type (& attributes)** | Absolute number of annotated items per annotation type (and associated attributes) | |

| **Term normalization (Grounding)** | Annotation types/instances with mappings into common (medical) terminologies, ontologies, lexicons | An example:<br>*ICD 10*: Disease$_{type}$ – bacterial pneumonia$_{instance}$<br>→ J15$_{ICD_10}$<br>*SNOMED CT*: Disease$_{type}$ – Tuberculosis (disorder)$_{instance}$<br>→ 56717001$_{SNOMED_CT}$ |
|---|---|---|
| **# Annotators + Mediators** | Absolute number of annotators (incl. educational background) & mediators/managers (incl. educational background) | |
| **IAA / Annotation type** | Scores for inter-annotator agreement (IAA) per annotation type under different matching conditions, if any, e.g.,<br>• strict match<br>• sloppy match | Comment: using metrics such as F1, Krippendorff's α, Cohen's κ, etc. |
| Average annotation time | Arithmetic *mean* (incl. standard deviation) or *median* of the time required to annotate all single metadata items per annotation type (for each annotator & aggregated for the entire annotation team, i.e., micro & macro IAA) | |
| Technical format | Storage format of the annotations | BRAT/BioC, JSON etc. |

**Table 7:** Corpus Card – a Template Datasheet for Corpus Descriptions, with Mandatory (bold) and Optional Attributes (non-bold)

This template primarily focuses on (clinically/medically relevant) content issues only. Provenance and legacy, storage and format requirements or legal/ethical issues will have to be added for a more comprehensive template (see, e.g., [111,112]).